\documentclass{article}

\usepackage[final]{corl_2020} 
\usepackage{graphicx, nicefrac}
\usepackage{wrapfig, booktabs}
\usepackage{amsmath}
\usepackage[ruled,vlined, linesnumbered]{algorithm2e}
\usepackage{multicol}
\usepackage{caption}
\usepackage[compact]{titlesec}

\title{STReSSD: Sim-To-Real from Sound \\for Stochastic Dynamics}

\author{
Carolyn Matl$^{1,2}$, Yashraj Narang$^{2}$, Dieter Fox$^{2,3}$, Ruzena Bajcsy$^{1}$, Fabio Ramos$^{2,4}$ \\
University of California, Berkeley$^1$, NVIDIA$^2$,
University of Washington$^3$, University of Sydney$^4$\\
\texttt{\{cmatl,bajcsy\}@eecs.berkeley.edu}, \texttt{\{ynarang,dieterf,ftozetoramos\}@nvidia.com}
}

\begin{document}
\maketitle

\begin{abstract}
    Sound is an information-rich medium that captures dynamic physical events. 
    This work presents STReSSD, a framework that uses sound to bridge the simulation-to-reality gap for stochastic dynamics, demonstrated for the canonical case of a bouncing ball.
    A physically-motivated noise model is presented to capture stochastic behavior of the balls upon collision with the environment. 
    A likelihood-free Bayesian inference framework is used to infer the parameters of the noise model, as well as a material property called the coefficient of restitution, from audio observations. 
    The same inference framework and the calibrated stochastic simulator are then used to learn a probabilistic model of ball dynamics.
    The predictive capabilities of the dynamics model are tested in two robotic experiments. First, open-loop predictions anticipate probabilistic success of bouncing a ball into a cup. The second experiment integrates audio perception with a robotic arm to track and deflect a bouncing ball in real-time. We envision that this work is a step towards integrating audio-based inference for dynamic robotic tasks. Experimental results can be viewed at \url{https://youtu.be/b7pOrgZrArk}.
\end{abstract}


\begin{figure}[h]
    \centering
    \includegraphics[width=\textwidth]{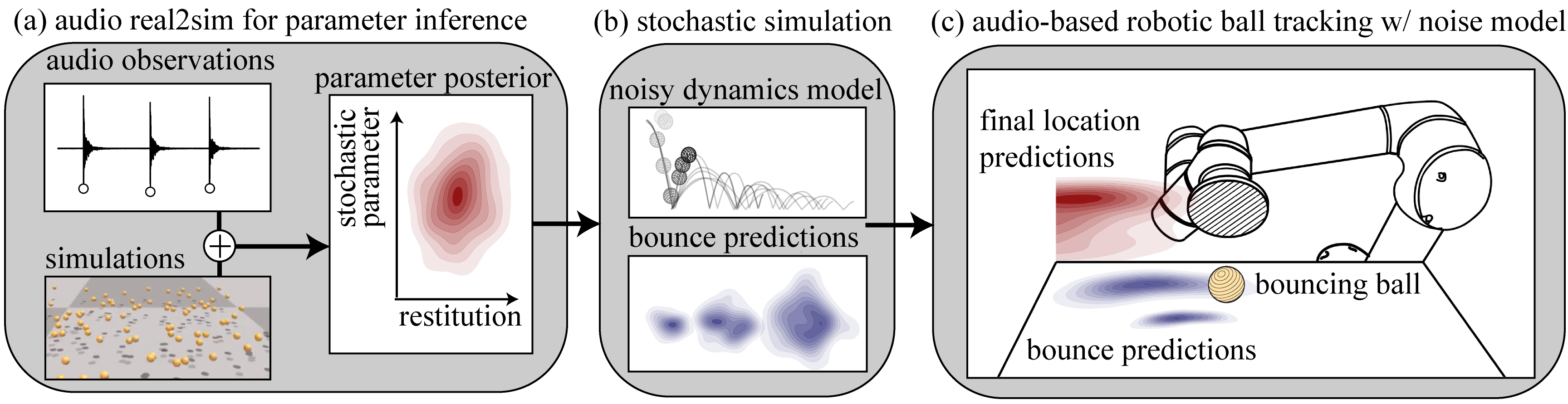}
    \caption{STReSSD is a framework that uses sound to bridge the sim-to-real gap for a stochastic dynamic process (e.g., a bouncing ball).
    (a) Simulation is used to learn a conditional density function relating a material and noise parameter to collision times and positions, and posteriors over these parameters are generated from audio observations of bouncing balls. (b) A probabilistic dynamics model is learned via the calibrated stochastic simulator. (c) This model is used with real-world auditory feedback for a reactive robotic ball-tracking task.}
    \label{fig:summary}
    \end{figure}

\section{Introduction}

The sensory modality of sound can impart practical physical intuition of the surrounding world. In particular, sounds arising from the dynamic interaction of objects or substances through direct contact are highly informative. These sounding interactions enable humans to accomplish various system identification tasks, such as discerning the size \citep{grassi2005we}, length \citep{carello1998perception}, and material \citep{klatzky2000perception} of a struck object. The abundance of information encoded in event-driven sound alone calls for the development of robotic techniques to extract physical understanding from this evidently descriptive modality. 

Thus, auditory perception presents a natural sensory choice to infer properties of and track an object that is dynamically interacting with its environment. State-of-the-art work has accomplished impressive results in dynamic object-tracking with visual input \citep{cigliano2015robotic, gomez2019reliable, li2012ping}. However, visual techniques are known to be sensitive to occlusions, specularities, transparencies, and poor lighting conditions. This work serves to complement visual methods with audio perception, which is robust to all these challenges. Furthermore, sound intrinsically is a lower-dimensional and higher-frame-rate signal, which consequently makes it much more computationally suitable for real-time robotic systems. 

A canonical dynamical system that is of interest to both the physics and robotics communities is the bouncing ball. Indeed, a wealth of prior work in the physics community has utilized impact sounds to estimate a ball's coefficient of restitution, a parameter that abstracts the energy dissipation upon collision \citep{bernstein1977listening, stensgaard2001listening, heckel2016can}. Nevertheless, it is widely acknowledged that modeling a ball's dynamics with a constant estimate of restitution does not realistically capture its stochastic behavior while interacting with the environment. Notably, \citet{heckel2016can} remarks that ``even tiny deviations of the shape from the perfect sphere may lead to substantial errors,'' specifically, due to directional perturbations upon collision (e.g., a tennis ball dropped on asphalt will likely veer off to a side). Prior work has not addressed this stochastic behavior, yet there is evidently a need for modeling these perturbations.

In robotics, simulations are often used to train robotic tasks with the assumption that the simulator models realistic dynamics, yet simulations are inherently deterministic. 
However, to capture realistic stochasticity within simulation, three challenges come to mind: 1) How does one automatically infer these stochastic parameters? 2) How can they be incorporated into a simulator? and 3) How can this now-stochastic simulator be utilized in a practical way for real robotic tasks?

\textbf{Contributions:}
As more research within robotics seeks to bridge the gap between simulation and reality, this work contributes to the community in three specific ways (Figure \ref{fig:summary}). First, we present a methodology to automatically calibrate simulators from simple features of sound. Second, we introduce learned stochasticity into a robotics simulator, in which a physics-inspired noise model accounts for real-world distributions observed in a bouncing ball's dynamics, e.g., due to uneven or rough surfaces of the ground or ball. Finally, we conduct experiments to demonstrate the predictability of the calibrated simulator in new scenarios, as well as the use of auditory feedback to achieve reactive control in a robotic ball-tracking task. This work combines techniques from multiple fields so that the resulting contributions may enhance the perceptual intelligence of robots using sound.

\section{Related Work}
	
	\subsection{Auditory Perception for Robotic Manipulation}
		Despite its ubiquity and low cost of capture and processing, auditory perception is not often utilized in real-time robotic manipulation. Instead, it has been most widely used for sound localization \citep{murray2009robotic,lang2003providing} and object or material classification \citep{brooks2005vibration, christie2016acoustics, krotkov1997robotic,  kroemer2011learning, sterling2018isnn, sinapov2014learning, chen2016learning}. However, only few works have successfully employed it for manipulation tasks. 
\citet{gandhi2020swoosh} showed that sound can be informative of actions applied to sounding objects and thus can be used to predict forward models of the objects.
	\citet{liang2019making} and \citet{clarke2018learning} also demonstrate dynamic inference from sound by using it for feedback control while pouring liquids and grains, respectively. In contrast, in this work, sound is additionally used to learn physically-relevant parameters, which then, combined with sound as feedback, inform fast robotic interactions with a bouncing ball. 
	
   \subsection{Simulation Calibration of Dynamics Parameters}
    
    A growing field in robotics lies in estimating the simulation-to-reality gap. Domain randomization is a popular technique used to learn policies in simulation that are transferable to the real world \cite{peng2018sim, chebotar2019closing}. \citet{ramos2019bayessim} develops BayesSim, a likelihood-free Bayesian approach to generate mixture-of-Gaussian posteriors over simulation parameters that are used for domain randomization. In general, BayesSim is more sample-efficient and generates better approximations of the posteriors than other likelihood-free inference techniques such as approximate Bayesian computation methods \cite{papamakarios2016fast}.  Recently, BayesSim has been successfully applied in different robotic contexts -- for instance, to calibrate granular material simulations, which can be used to predict real-world behavior \citep{matl2020inferring}.

This work leverages BayesSim to approximate a posterior over a material and noise parameter, which is sampled for domain randomization when simulating a bouncing ball.
Both \citep{allevato2019tunenet} and \citep{asenov2019vid2param} use real2sim methods on video input to infer bouncing ball parameters. 
Our work is unique in that sound is used in lieu of vision, and a noise model is learned to account for observed stochasticity. 
Learning stochastic parameters has been shown to be useful in works such as \cite{bauza2017probabilistic, bauza2018data, ma2018friction}, where probabilistic models are used to make predictions of planar pushing with frictional contact.
The only known works other than this paper that use sound for real2sim focus on realistic sound synthesis \citep{zhang2017shape}. 
However, \citet{zhang2017shape} acknowledge their limitations in sim2real transfer, as they rely on bridging the gap using high-dimensional spectral features of sound. In contrast, this paper extracts simple features from audio, making it robust to discrepancies in vibration representations in simulation and real life as well as more suitable for real-time applications. 
   
   \subsection{Robotic Bouncing Ball Tracking}
   
   A bouncing ball is difficult to track due to uncertainties in collision dynamics.
However, robot interactions with a dynamic object require accurate tracking and trajectory predictions. 
Several works have attempted bouncing-ball tracking with vision, either by using high-speed or multiple cameras \citep{gomez2019reliable} or pairing low-cost cameras with sim2real tuning of physically-relevant parameters \citep{allevato2019tunenet, asenov2019vid2param}. By using dynamics models trained in the tuned simulator, trajectory predictions compensate for tracking uncertainties \cite{asenov2019vid2param}. In contrast, this work presents ball tracking through sound and physical parameter inference. The only known work that uses sound to interact with a bouncing ball is \citep{kuhn_2018}, where the author designed a mechanism that continuously bounced a ball off an actuated platform. However, sound is used reactively, rather than to inform predictions about the ball dynamics.

\section{Methodology}
\label{sec:framework}

The STReSSD framework combines BayesSim \citep{ramos2019bayessim} with audio processing to calibrate a stochastic dynamic simulator. We highly recommend referring to Appendix \ref{app:flowdiagram}, which illustrates an expanded version of the STReSSD framework. The following sections detail each module in the framework.

\subsection{Simulation Calibration with BayesSim} 
	
		This work aims to calibrate a physics simulator using auditory signals of ball-table collisions to closely predict real-world dynamics of the ball. Let $\theta$ be the simulation parameters to be inferred, and let $X$ represent information-rich features of an auditory signal. Assuming that there is an equivalent representation of $X$ in simulation ($X^s$) and the real physical environment $(X^r)$, then the problem can be formulated as computing the posterior $p(\theta | X^r)$ over the simulation parameters. This work uses a likelihood-free framework called BayesSim \cite{ramos2019bayessim} to approximate this posterior function. Specifically, $N$ pairs of parameters and feature vectors $\{ \theta_i, X^s_i\}^N_{i=1}$ are first generated from forward simulations of samples from a physically-motivated prior distribution, $p(\theta)$. A conditional density function $q_\Phi(\theta | X^s)$ is then learned by mapping extracted simulation features $X^s$ to a mixture of $K$ Gaussian components with a mixture density neural network (MDNN), parameterized by $\Phi$. The conditional density function $q_\Phi$, along with a real observation $X^r$, are used to generate an approximation of the posterior $p(\theta|X^r)$. The main advantage of using BayesSim over non-Bayesian techniques (e.g., classical optimization) is that the approximated posterior can be informative of the uncertainties regarding the parameters $\theta$, as well as sensitivity to measurement noise.

    \subsection{Stochastic Simulation of a Bouncing Ball}
    \label{sec:stochasticsimulation}
	
	\begin{figure}[h]
    \centering
    \includegraphics[width=\textwidth]{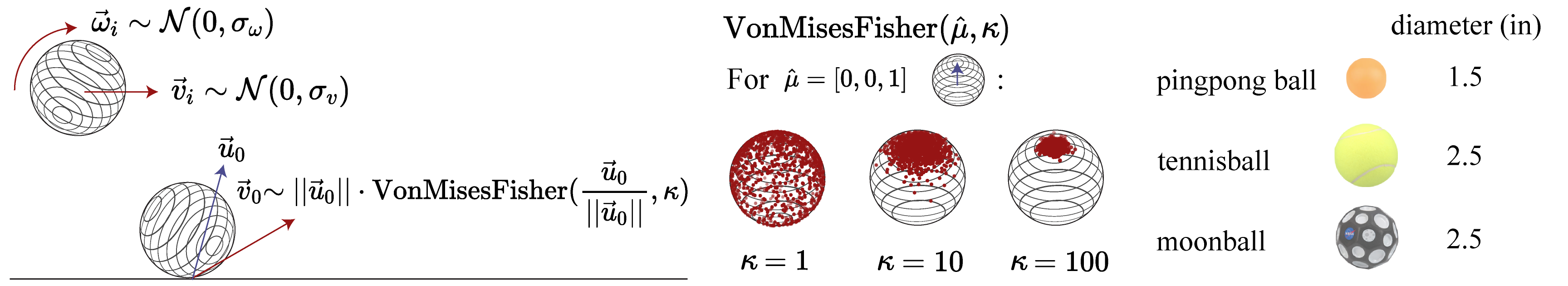}
    \caption{(Left): The initial velocity is sampled from a Gaussian distribution and collision perturbations are sampled from a von Mises-Fisher distribution, centered at $\nicefrac{\vec{u}_0}{||\vec{u}_0||}$. (Middle): 1000 samples over a sphere for different values of $\kappa$ for $\mu=[0,0,1]$. (Right): Balls used in real experiments.}
    \label{fig:noisemodel}
    \end{figure}
    
    Here, we define the simulation parameters $\theta$ that are used to model the stochastic dynamics of a bouncing ball. 
    The coefficient of restitution $e$ is a standard ratio used to quantify energy loss upon object collision. For an object bouncing on an immovable surface, $e$ can be calculated as the ratio of the exiting to entering velocity of the collision. However, as velocities are difficult to extract from sound, bounce times are used to calculate $e = \frac{\text{time from bounce $i$ to $i+1$}}{\text{time from bounce $i-1$ to $i$}}$ (see Appendix \ref{sec:derivationofe}). 

    While $e$ is generally used as a \textit{constant} in simulation, numerous factors contribute to nontrivial variance in measurements of $e$ \citep{montaine2011coefficient}. For example, if a ball is dropped vertically onto a level flat surface, surface asperities cause the ball to experience horizontal perturbations. Because we are measuring $e$ from bounce times instead of velocity, these horizontal perturbations produce a \textit{distribution} over $e$.
  
    \citet{montaine2011coefficient} address this variance by fitting a probability density over $e$. 
    Instead, this paper aims to represent $e$ and surface asperities as independent variables. To simulate collision perturbations, \citet{montaine2011coefficient} represent a ball as a composite multisphere particle with $\sim10^6$ surface asperities. 
    In contrast, this paper reduces computational complexity by imitating the stochastic dynamics of a bouncing ball with a simple noise model.
    At every collision, we apply a random perturbation, sampled from a Gaussian distribution on a sphere (i.e., the von Mises-Fisher distribution), to the direction of the outgoing velocity vector. 
    The von Mises–Fisher distribution on a 2-sphere (Figure \ref{fig:noisemodel}) is given by the probability density $f(x; \mu, \kappa) = \frac{\kappa}{2\pi (e^\kappa - e^{-\kappa})}e^{(\kappa \mu^T x)}$ for $3$-dimensional random unit vector $x$, where $\mu$ is the mean direction and $\kappa$ is a concentration parameter.
    
    Let $\vec{u}_0$ represent a ball's unperturbed exit velocity during a collision. To simulate a collision perturbation (e.g., due to rough surfaces), a vector is sampled from $f(x; \hat{\mu}, \kappa)$, where $\hat{\mu}=\frac{\vec{u}_0}{||\vec{u}_0||}$. The vector is then scaled by $||\vec{u}_0||$ to produce a perturbed exit velocity $\vec{v}_0$. Thus, the outgoing kinetic energy of the ball is unaffected by the perturbation. Stochastic ball dynamics are therefore modeled by $e$, for energy dissipation, and $\kappa$, to account for the stochasticity due to collisions.
    
    While this model can be integrated with most standard simulators, we used NVIDIA's Isaac Simulator paired with a preconditioned conjugate residual (PCR) solver \cite{macklin2019non}.
    This GPU-optimized simulation platform enabled the parallelization of hundreds of environments. This was
    particularly useful for efficiently generating distributions of ball trajectories due to the modeled collision perturbations.

	\subsection{Audio Processing}
	
	\begin{figure}[h]
    \centering
    \includegraphics[width=\textwidth]{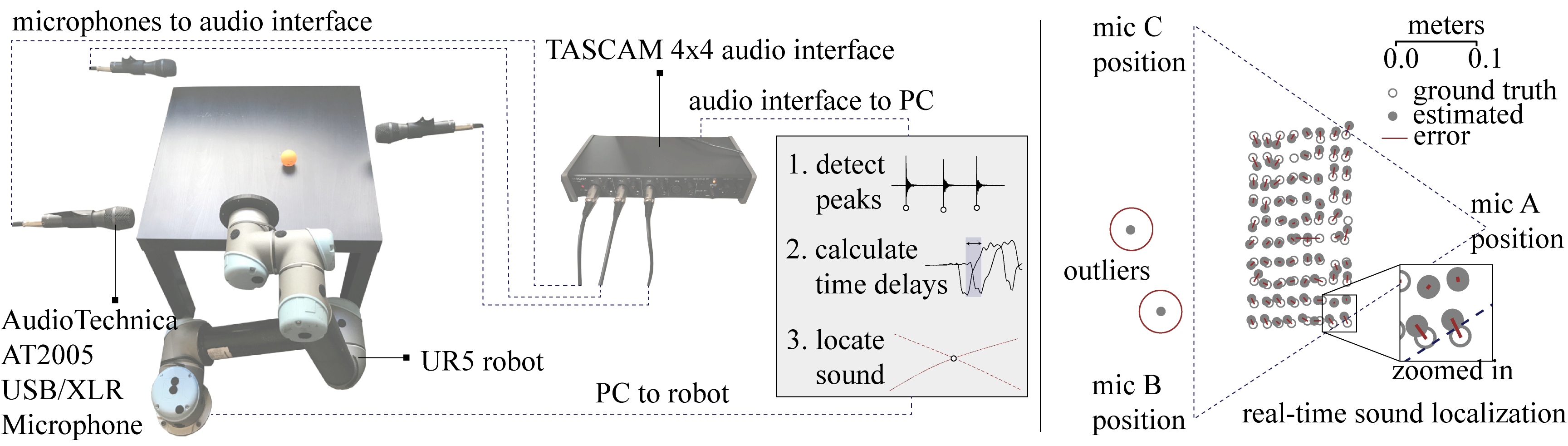}
    \caption{(Left): Integrated robotic system used in experiments. (Right): Real-time sound localization is slightly less accurate than the offline method and produces outliers with large errors.}
    \label{fig:acousticsetup}
    \end{figure}

    We hypothesized that the times and positions of the ball bounces are informative of $\theta=\{e,\kappa\}$, as $e$ can be expressed as a function of impact times (Section \ref{sec:stochasticsimulation}), and $\kappa$ affects the ball bounce positions. This motivates our four-dimensional feature vector $X$, which encodes both temporal and positional information. Letting $t_i$, $\vec{p}_i$, and $d_i$ denote the time, positional vector, and distance between bounce $i$ and $i+1$, then $X$ is composed of $\frac{t_2}{t_1}$, $d_1=||\vec{p}_1||_2$, $d_2=||\vec{p}_2||_2$, and $\alpha$, the signed angle between $\vec{p}_1$ and $\vec{p}_2$ (see Appendix \ref{app:featurevector}). Higher dimensional spectral features were not needed to estimate $\theta$, but future work could leverage this additional information to, for instance, classify the ball's material.
    
    Vector $X$ is simple to extract from simulation, but, in the real-world, depends on accurate sound localization for each bounce. We used two localization methods -- an offline method for parameter inference and an online method for real-time interaction, both of which relied on the same principle of time delay estimation (TDE). TDE uses time delays between microphone signals to locate sound sources (see Appendix \ref{app:soundlocalization} for derivation). For 2D localization, only 3 microphones are necessary. 

    	The offline and real-time methods diverged in how these time delays were calculated. Offline time delays were calculated using phase correlation, a standard frequency-domain technique \cite{kuglin1975phase}, over audio segments of $>$20 ms. In contrast, real-time processing required fast calculations on 11 ms-length buffers of incoming audio.  Peak times corresponding to the strong peaks induced by collisions could be detected efficiently and thus were used to calculate time delays. The offline method was found to be slightly more accurate, with an average error of about 6.7 mm compared to real-time's 7.9 mm, shown in Figure \ref{fig:acousticsetup} (see Appendix \ref{app:soundlocalization} for comparisons). Experiments were conducted in a household environment rather than a sound studio, so real-time localization was slightly less robust. The method's sensitivity to reverberation, or echoes from early reflections off of walls and furniture, occasionally caused large errors in detecting true impact times. However, the performance was a necessary trade-off to enable reactive robot interactions. Furthermore, peak detection was generally unaffected by other factors such as noisy robot motion or acoustic differences in ball material.

For experiments, auditory signals were captured by three AudioTechnica AT2005 USB/XLR Microphones, which were placed around a 0.55-by-0.55 meter particle-board table, 3 inches above the table surface. Microphone signals were streamed at 44.1 kHz in parallel to a PC through a TASCAM 4x4 audio interface (Figure \ref{fig:acousticsetup}). For robot experiments, audio was processed in real-time, and control signals were sent to a UR5 robot in parallel processes. The overall latency of the system was dominated by robot motion, as bounce localization and inference ran at $\sim$100 and 20 Hz, respectively.

	\subsection{Robotic Ball Tracking}\label{sec:roboticballtracking}
\begin{figure}[h]
    \centering
    \includegraphics[width=\textwidth]{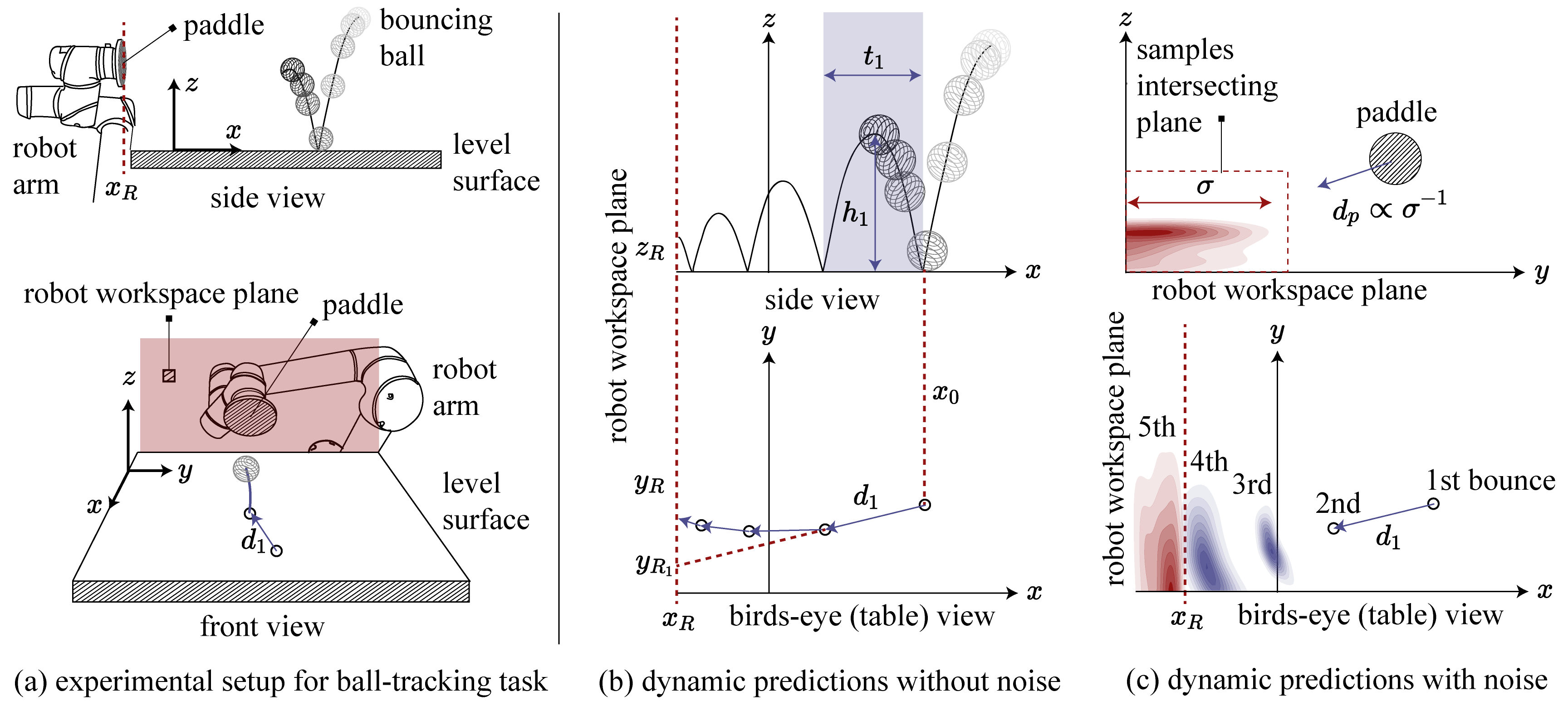}
    \caption{(a) Ball tracking setup. (b) Example ball trajectory for noiseless method. $y_{R_1}$ is the $y$-coordinate of the inferred crossing point of the ball given the first two bounces. (c) Samples of future bounce positions given the first two bounces using the stochastic noise method. The mean and variance of the samples in the workspace plane inform the distance the paddle should travel ($d_p$).}
    \label{fig:roboticballtracking}
    \end{figure}
	
	To illustrate the benefits of modeling dynamic stochasticity and the real-time capabilities of audio perception, we compare two methods to accomplish the robotic task of tracking and deflecting a bouncing ball using sound. The methods differ in their consideration of perturbations upon collision. Pseudocode and equations accompanying each method are presented in Appendices \ref{app:noiseless} and \ref{app:noisy}.

	\subsubsection{Deterministic Baseline: Dynamic Predictions without Collision Noise}
	\label{sec:dynamic_predictions_nonoise}
	
		Without modeling collision perturbations (thus ignoring concentration parameter $\kappa$), the dynamics of a bouncing ball are defined only by its initial conditions, gravity, and energy-dissipating forces. For this particular method, we assume that the most significant parameter affecting ball dynamics is $e$. The ball trajectory can then be defined by classical projectile equations, with $e$ applied to simulate energy dissipation at each bounce. Figure \ref{fig:roboticballtracking}a-b illustrate the control scheme used to guide the robot to contact the ball as it enters the workspace plane. 
	The ball's future bounces can be extrapolated from the times and positions of the last two bounces, since here we assume that the ball is not directionally perturbed. The goal position of the robot, denoted as $(x_R, y_R, z_R)$ in Figure \ref{fig:roboticballtracking}, is then defined as the intersection of the piecewise parabolic trajectories with the robot plane. If the ball will intersect the plane at $t_R$ seconds before the next bounce, then the robot moves to the goal position in that time. Otherwise, the robot moves to $y_R$ before the next bounce is observed.

	\subsubsection{Stochastic Method: Dynamic Predictions with Collision Noise}
	\label{sec:dynamicpredictions_collisionnoise}
	 Predicting the dynamics of a ball without incorporating knowledge of collision noise may cause the robot to be over-reactive to small perturbations in the ball's trajectory. Furthermore, success relies on accurate real-time sound localization, which as mentioned earlier, can occasionally be sensitive to reverberations, causing large errors in the inferred bounce location (Figure \ref{fig:acousticsetup}). Thus, we aim to incorporate the noise model to address two goals: leveraging the uncertainty of the ball's location to prevent the robot from over-reacting, and filtering unrealistic inferred bounce locations.

Both goals employ a second BayesSim model that generates posteriors over the next bounce time and location, given the time and distance between the prior two bounces. The BayesSim model, trained via the calibrated simulator, serves as a probabilistic dynamics model. At run-time, a generated posterior is used with an elliptic envelope filter to classify if the observed bounce can be trusted (or if it is an outlier). If the latter, the mean of the posterior is used to estimate the bounce location. 

In addition, the ball's future trajectory is estimated by generating samples from the posterior, which in turn are used to generate new posteriors. Thus, trajectory predictions are made by recursively generating posteriors until samples intersect the plane. In this new context, BayesSim is used as a generic regression technique to learn a transition model, which is applied recursively, and whose predictions are averaged. This dynamics model
could also be incorporated into a nonlinear filtering technique such as an Unscented Kalman Filter, although this would require defining a sensor model.
Figure \ref{fig:roboticballtracking}c shows that the predicted future bounce location uncertainty grows with every bounce. This uncertainty informs the end-effector motion. The smaller the variance of the predictions are, the closer the end-effector will move toward the goal location before the next observation.

\section{Experiments}
\label{sec:result}

    \begin{figure}[htp]
    \centering
    \includegraphics[width=\textwidth]{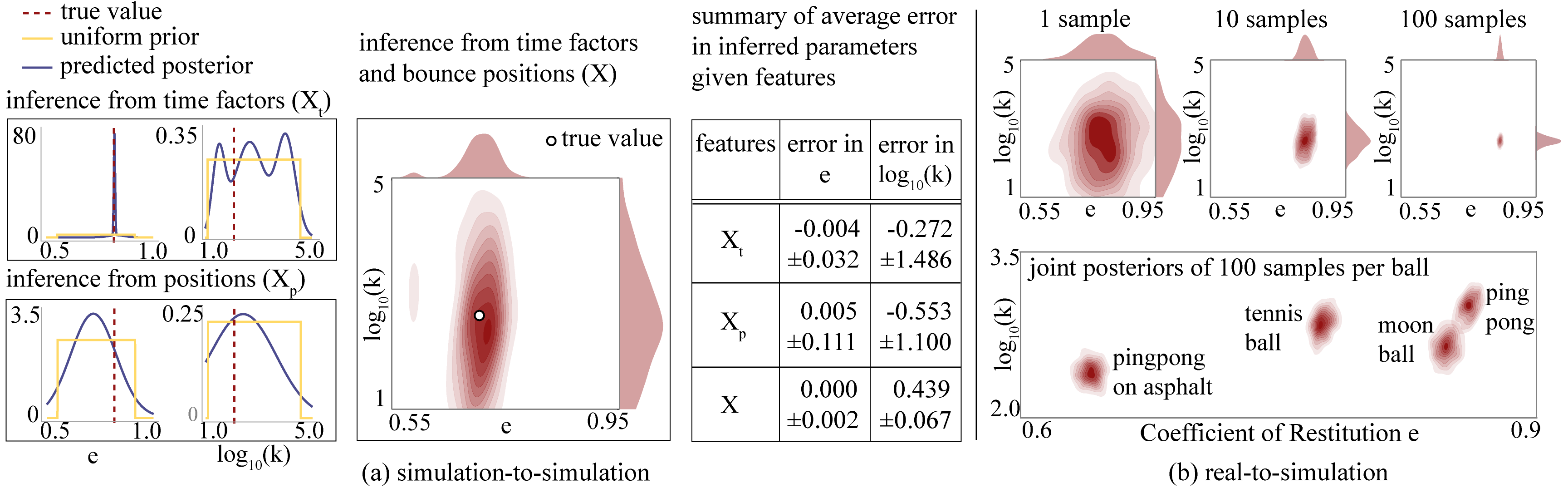}
    \caption{(a) Sim-to-sim example posteriors of inferring $e$ and $\log_{10}(\kappa)$ from bounce times,  $e$  and $\log_{10}(\kappa)$ from bounce positions, and both $e$ and $\log_{10}(\kappa)$ from times and positions. The table summarizes average error over 1000 examples for each parameter, given a feature subset. (b) Top: Real-to-sim posteriors become more certain with more samples. Bottom: Posteriors for real balls.}
    \label{fig:simulationresults}
    \end{figure}

	\subsection{Simulation-to-Simulation Experiments}
	The inference framework was first tested with simulated data to assess whether bounce locations and times were relevant measurements for the inference of the desired model variables ($e$ and $\kappa$). Training data was generated in simulation by sampling from a physically-motivated uniform prior distribution, with $e \in [0.55, 0.95]$ and $\log_{10}(\kappa) \in [1,5]$. Each simulation sampled an initial angular and linear velocity from a Gaussian centered at $0$ (see Figure \ref{fig:noisemodel}) to mimic real-world noise when dropping a ball. 6000 and 1000 simulations were run for the training and testing dataset, respectively.

To measure how informative each feature in $X$ was for each parameter in $\theta$, an ablation study was performed over the simulated dataset. The features were divided into two sets, one relevant to time ($X_t=\frac{t_2}{t_1}$), and the other, to position ($X_p=\{d_1, d_2, \alpha\}$). Four BayesSim models were trained, each using either $X_t$ or $X_p$ to infer one of $e$ or $\kappa$. These models were used to generate posteriors from a testing dataset of 1000 examples, and the peaks of the posteriors were compared with the true value of the inferred parameter.  The table in Figure \ref{fig:simulationresults}a summarizes the average error in inferred parameters $e$ and $\kappa$, given a subset of features. Posteriors that are unimodal and low-variance have higher certainty and thus relevance to the inferred parameter. As reflected in the single-parameter posteriors of Figure \ref{fig:simulationresults}a, time appears to be a more informative observation for $e$, while positional information is more relevant for $\kappa$. Knowledge of both time and positional information generates significantly more certain posteriors. Observe that the variance of the marginal posterior for $\log_{10}(\kappa)$ will naturally be greater than that of $e$, as estimating a noise parameter is inherently more difficult.

	\subsection{Real-to-Simulation Experiments}
	\label{real2sim}
	
	With the assurance that $X$ contained informative features for $\theta$, we turned to extracting $\theta$ to describe real bouncing balls. Note that both $e$ and $\kappa$ are reflective of material-to-material interactions, and thus are unique to each pair of object and surface. The four object-surface pairs tested were a ping-pong ball, tennis ball, and moon ball (which has large crater-like divots) bouncing on a smooth table, and a ping-pong ball bouncing on an asphalt surface (see Figure \ref{fig:noisemodel}). This was to demonstrate the effect of surface asperities both on the ground (asphalt) as well as on the ball (e.g., moon ball). 
	
	For each ball-surface pair, the ball was dropped 100 times from 0.26 meters (to match simulation), and offline sound localization captured the locations and times of the bounces.  Real-world feature vector $X^r$ was extracted for each drop and used to generate a posterior over $\theta$. Figure \ref{fig:simulationresults}b shows an example posterior generated from one sample, or drop, of the moon ball. Note that, from one sample, the posterior is quite uncertain. However, assuming independence between each new observation $X^r_i$, the joint posterior can be generated by multiplying each individual posterior together. As the single sample posteriors reflected strongly Gaussian structures, each mixture-of-Gaussians was projected to a single Gaussian before multiplication. As shown in Figure \ref{fig:simulationresults}b, the joint posterior became more certain with more observations, with the joint posterior of 100 samples demonstrating strongly peaked marginals in both $e$ and $\log_{10}(\kappa)$. 	The resulting posteriors of each ball-surface pair are illustrated in Figure \ref{fig:simulationresults}b, and their positions relative to each other qualitatively reflect their real-world dynamic characteristics (e.g., the moon ball is bouncier and noisier than the tennis ball). 

	\subsection{Closing the Loop: Real-to-Sim-to-Real}
	
	\begin{figure}[h]
    \centering
    \includegraphics[width=\textwidth]{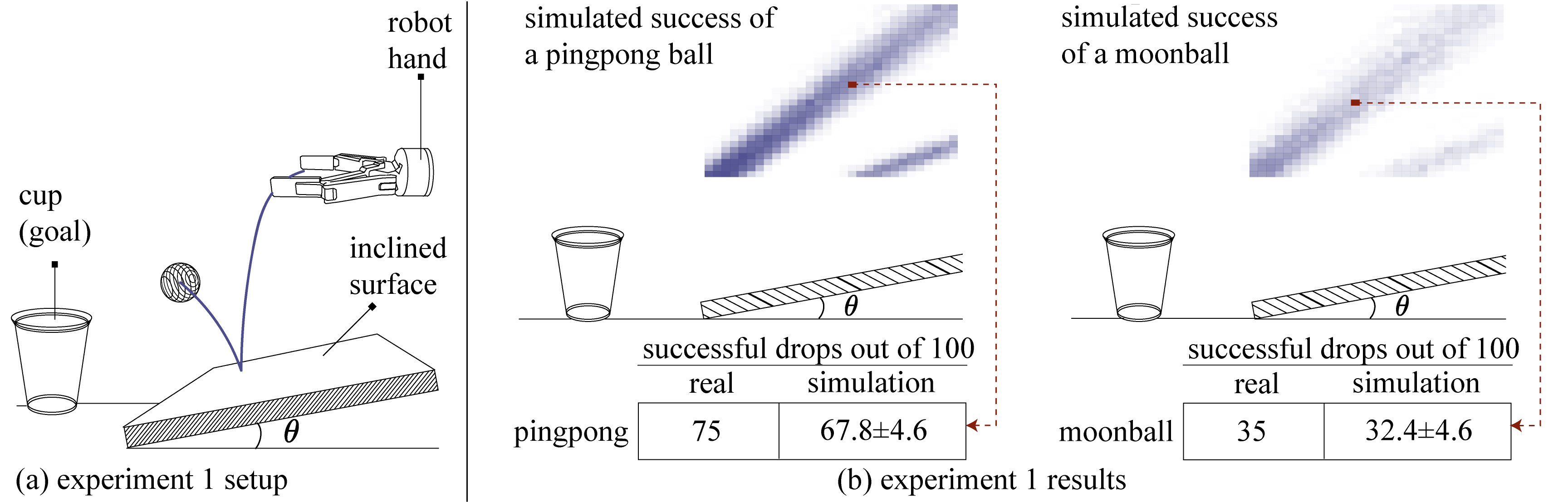}
    \caption{(a) A robotic hand bounces the ball off the tilted table surface into the cup, or goal. (b) Top: Spatial map of success rates for different initial positions of the ball drop. Success ranges from lightest (0 successful) to darkest (100 successful). Bottom: Comparison of real and simulated success rates (out of 100) dropped at the red point, with simulated success rates averaged over 30 trials of 100 simulations. }
    \label{fig:experiment1}
    \end{figure}
    
    In order to test if the dynamic model of the bouncing ball generalized well to new scenarios, an experimental setup similar to the one constructed in \citet{allevato2019tunenet} was used, as illustrated in Figure \ref{fig:experiment1}. The table surface used in Section \ref{real2sim} was tilted at an angle of $10.72 \deg$, and a plastic cup was placed 8.6 inches away from the bottom of the incline. The physical experimental setup was replicated in simulation, and the goal was to compare the success rate of bouncing a ball against the inclined plane into the cup in simulation and in real experiments. While a deterministic simulator would only either fail or succeed if a ball was dropped at a specific initial position, by modeling dynamic noise due to surface perturbations, our stochastic simulator had probabilistic success.
    
     Using a Robotiq 2F-140 gripper to drop the ball repeatedly over the inclined surface at a fixed position, real experiments showed similar probabilistic levels of success, with the pingpong ball and moon ball bouncing into the cup (out of 100) 75 and 35 times, respectively. The experiment was repeated in simulation 30 times, for each ball and their respective drop positions.  For each simulation, a new pair of $e$ and $\log_{10}(\kappa)$ were sampled from the inferred posteriors from Section \ref{real2sim}. The average simulated success rates were within close range of the real success rates, suggesting that the learned dynamics model could generalize well to novel physical scenarios. This also implies that the simulator could potentially be used for predictive measures, e.g., searching for where to drop a ball to maximize success rate. To demonstrate this concept, we sampled success rate (out of 100) over a discrete number of initial positions in simulation. This generated a spatial map of probabilistic success (Figure \ref{fig:experiment1}), a capability made possible through the modeling of collision perturbations.
	
	\subsection{Robotic Ball-Tracking Results}
	\label{sec:roboticballtrackingresults}
	\begin{figure}[h]
    \centering
    \includegraphics[width=\textwidth]{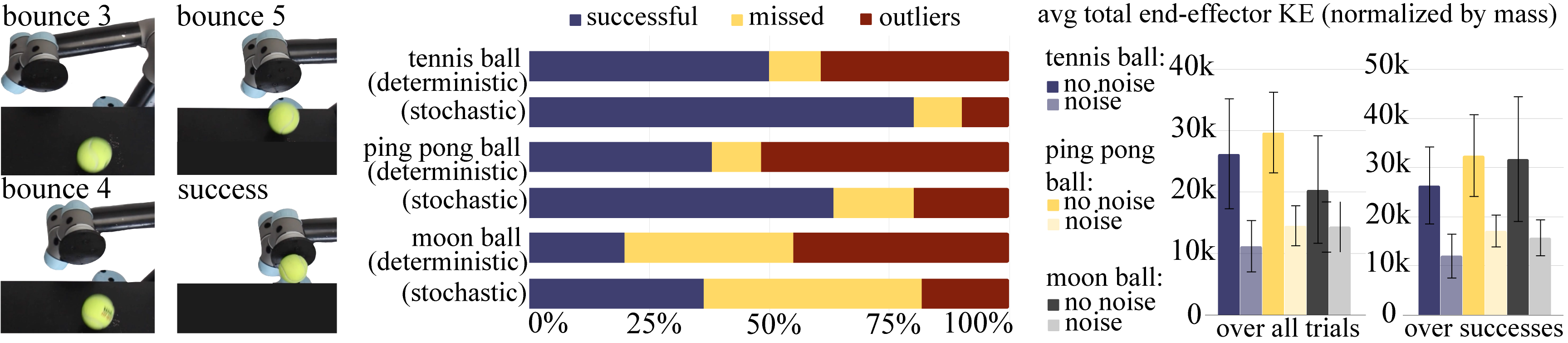}
    \caption{(Left): A successful trial. (Middle) ``Outliers" signifies failures due to sound localization errors,  and ``Missed'' denotes any other failure. Compared to the deterministic baseline, the stochastic method has higher success (an absolute increase of 24\%) and fewer failures due to outliers (an absolute decrease of 29\%), averaged across all balls. This graph excludes trials that violated the robot's boundary constraints; an alternative graph with all 30 trials can be found in Appendix \ref{sec:appendixd1}. (Right): On average, the baseline uses more total energy than the stochastic method over all trials and all successful trials, suggesting that the baseline is more overreactive.}
    \label{fig:experiment2}
    \end{figure}
    
    The goal of the ball-tracking task was to move the robot end-effector, a circular paddle with a diameter of 4.5 inches, to touch the bouncing ball as it entered the robot workspace plane, using only sound as feedback. Each of the methods described in Section \ref{sec:roboticballtracking} was evaluated for the tennis ball, ping-pong ball, and moon ball on the following metrics: 1) success out of 30 trials (whether or not the end-effector touched the ball), and 2) kinetic energy of the end-effector, integrated over the trajectory. We hypothesized that, by modeling collision perturbations, the probabilistic method would be more robust to errors in bounce localization as well as curb the reactiveness of the robot to small perturbations in the ball's trajectory. Both methods assumed nothing about the initial condition of the ball, and for each trial, the ball was manually tossed onto the table in the direction of the robot.

     As shown in Figure \ref{fig:experiment2}, the stochastic method averaged an absolute increase in success rate of $24\%$, which corresponds to an improvement of $70\%$, relative to the deterministic baseline.
    Failure cases are labeled as ``outliers'' for errors in bounce localization and ``missed'' for all other failures. For each of the balls, a few of the 30 tosses resulted in the ball violating the robot workspace boundaries, so those tosses were removed from analysis. Sensitivity to ``outliers'' in bounce localization can be resolved by using dynamic predictions with collision noise described in Section \ref{sec:dynamicpredictions_collisionnoise}. Figure \ref{fig:experiment2} precisely demonstrates this effect, where the number of failure cases due to outliers in bounce localization diminished for all three balls. While the stochastic method eliminates failures due to outliers in later bounces, it is still sensitive to outliers in the first and second bounces, since it relies on those to generate posteriors for future bounces. Finally, Figure \ref{fig:experiment2} shows that the average total end-effector kinetic energy over all trials, normalized by mass and averaged across balls, was reduced by $46\%$ when considering collision noise versus without. This pattern is replicated when looking at the average energy over just the successful trials (a reduction by $51\%$), suggesting that the stochastic method does indeed curb over-reactions to small perturbations in a bouncing ball's trajectory.

\section{Conclusion}

This work presents STReSSD, a novel framework for calibrating a simulator for stochastic dynamics from simple, low-dimensional features of sound. A physically-motivated noise model is learned from real audio, which enables a physics simulator to reflect real stochastic dynamics caused by rough or uneven surfaces. Using the canonical dynamical system of a bouncing ball, this work closes the loop on real-to-sim-to-real by demonstrating the predictive capabilities of the learned dynamics model in both an open-loop and sound-in-the-loop robotics task. To the best of our knowledge, this is the first robotics work that relies solely on sound to accomplish a highly dynamic, probabilistic task. 
We believe that the robotic capabilities demonstrated by the devised framework highlight the utility of audio perception and may encourage increased integration of sound in robotics. 

While the experimental results are promising, there are various opportunities to extend this work. Sensor fusion with vision and more robust peak-detection can improve the accuracy of real-time sound localization. 
Furthermore, one could imagine utilizing the rich spectral information from the raw audio signals to classify the material of the ball. The multivariate classification uncertainty could then be incorporated into the BayesSim model to produce a multimodal posterior distribution over simulation parameters.
Future work aims to explore simulation-in-the-loop inference, where the simulator is calibrated in real-time and a dynamic object is tracked using a particle filter or an Unscented Kalman Filter. These advancements may lead to broader impact in various sub-fields of robotics, including navigation and unseen-object manipulation. 

\clearpage

\acknowledgments{The authors were supported in part by the National Science Foundation Graduate Research Fellowship. We thank Matthew Matl and all reviewers for their
insightful feedback and suggestions.}

\clearpage

\appendix
\section{STReSSD framework}
\label{app:flowdiagram}

\begin{figure}[h]
    \centering
    \includegraphics[width=\textwidth]{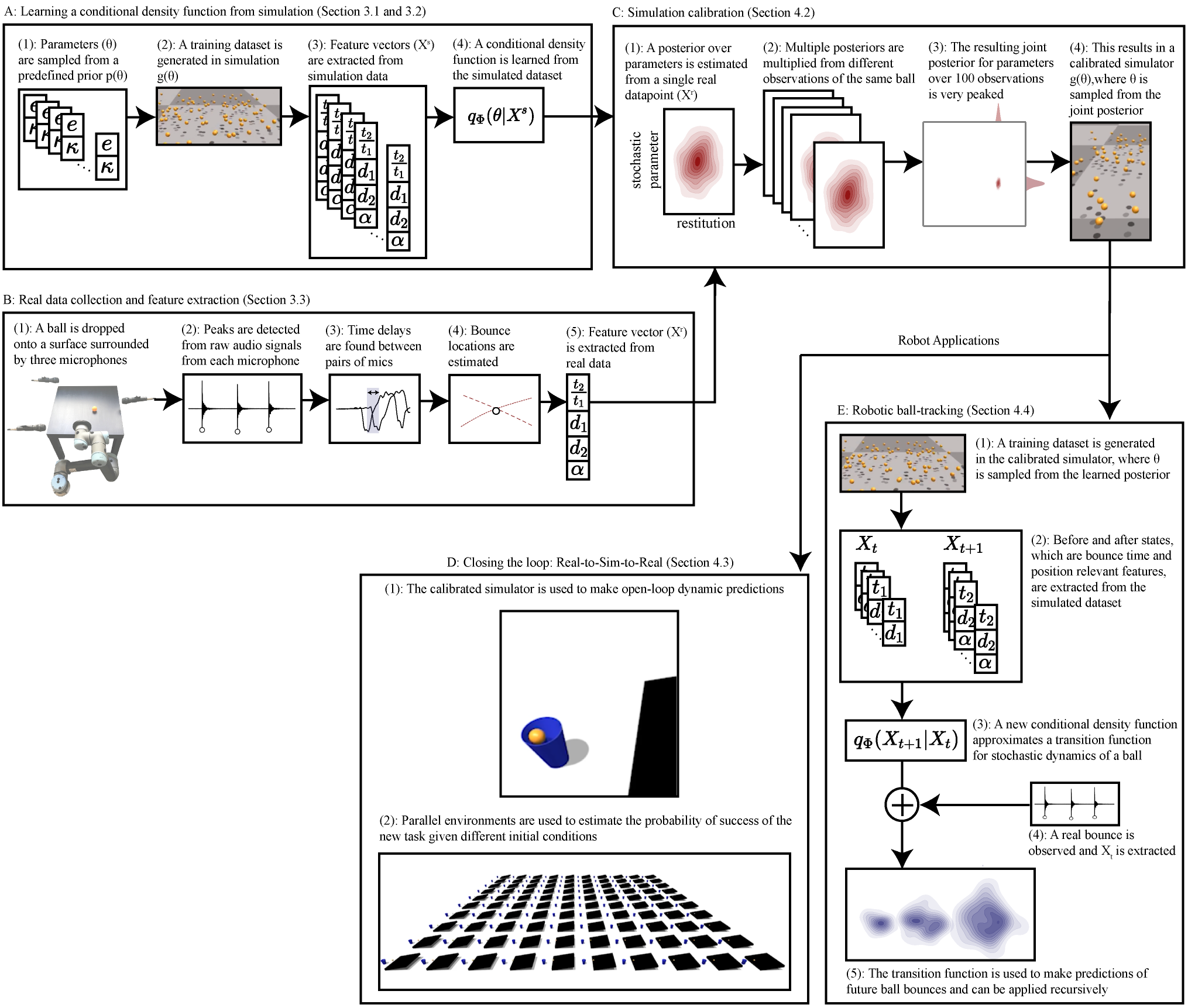}
    \caption{A flow diagram detailing the STReSSD framework. (A): Simulation is used to learn the conditional density function $q_\Phi(\theta|X^s)$ relating parameters $e$ and $\kappa$ to observation features $X=[\frac{t_2}{t_1}, d_1, d_2, \alpha]$. 6000 independent simulations are run to generate the training set to learn $q_\Phi$. (B): Feature extraction of real data begins by observing the peaks of raw audio signals, finding the time delays between pairs of microphones, and locating the bounces. The low-dimensional feature vector $X^r$ is thus extracted from raw data. (C): The simulator is calibrated using real observations to match a particular ball's dynamic behavior. A posterior over the parameters $\theta=[e,\kappa]$ is approximated using the learned conditional density function $q_\Phi$ and the real observation $X^r$. Posteriors from different observations of the same ball can be multiplied to generate a more peaked joint posterior. The calibrated simulator samples from this joint posterior. (D): The calibrated simulator from (C) is used to make open-loop dynamic predictions of the ball. (E): The calibrated simulator is used for robotic ball-tracking. A new training set is generated by forward simulating the calibrated simulator, with relevant parameters sampled from the joint posterior in (C). New features ($X_t = [t_t, d_t]$ and $X_{t+1} = [t_{t+1}, d_{t+1}, \alpha_{t+1}]$) are extracted from this simulator and a second conditional density function approximates a transition function for the stochastic dynamics of the ball. The transition function is used to estimate the next bounce location and time, given the last two bounces. The transition function can be applied recursively to predict future ball bounces.}
    \label{fig:flowdiagram}
    \end{figure}

\section{Feature vector $X$}
\label{app:featurevector}
Feature vector $X$ is composed of four elements. Let $X_t$ refer to the feature in $X$ relevant to time and $X_p$ refer to the three features relevant to position. Below, we derive the relationship between the coefficient of restitution $e$ and feature $X_t$, as well as how the features in $X_p$ are extracted from the positions of the first three ball bounces.

\begin{figure}[h]
    \centering
    \includegraphics[width=\textwidth]{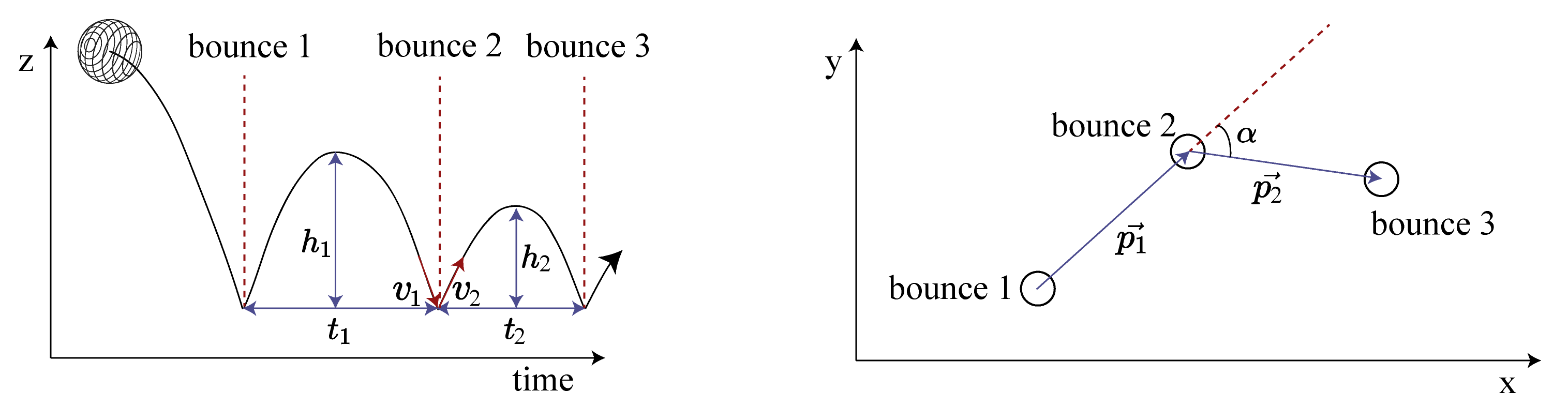}
    \caption{(Left): A graph of the vertical trajectory of a ball over time, labeled with relevant variables for extracting $X_t$. (Right): $xy$ planar positions of ball bounces are denoted by the open circles, and relevant variables for extracting $X_p$ are labeled.}
    \label{fig:appendixa}
    \end{figure}

\subsection{Derivation of $X_t$ from $e$}
\label{sec:derivationofe}
The coefficient of restitution $e$ is an expression relating the relative kinetic energy of a system before and after a collision. For a bouncing ball on a perfectly rigid surface, $e$ can be expressed by the following ratio: 
\begin{equation}
    e = \sqrt{\frac{\text{KE}_{\text{after collision}}}{\text{KE}_{\text{before collision}}}}
\end{equation}

We would like to extract $e$ from audio signals of a bouncing ball, assuming that it is dropped with $0$ initial linear velocity. With this assumption, the problem can be simplified to 1-dimension, where the ball travels only in the vertical direction (this is, of course, assuming for now that the ball is not perturbed upon collision). 
Letting $v_1$ and $v_2$ be the ball velocity before and after the 2nd bounce (see Figure \ref{fig:appendixa}), then $e$ can be rewritten in terms of the entering and exiting velocities:
\begin{equation}
    e = \sqrt{\frac{\nicefrac{1}{2}mv_2^2}{\nicefrac{1}{2}mv_1^2}} = \frac{v_2}{v_1}
\label{eq:2}
\end{equation}

However, velocities are difficult to measure or perceive with sound. Thus, we want to relate $e$ to the times between bounces 1 and 2 ($t_1$) and bounces 2 and 3 ($t_2$). Let $h_1$ and $h_2$ be defined by the heights depicted in Figure \ref{fig:appendixa}. Assuming that, between collisions, the bouncing ball trajectory can be defined by classical projectile equations (assuming an absence of factors such as air resistance), then $v_1$ can be related to $t_1$ by the following derivation:

We know from classical projectile equations that: 
\begin{equation}
    \frac{1}{2}mv_1^2 = mgh_1 \text{ and } \frac{1}{2}g(\frac{t_1}{2})^2 = h_1
\end{equation}

Then combining the above equations, we get a relationship for $v_1$ and $t_1$:
\begin{equation}
    \frac{1}{2} v_1^2 = \frac{1}{2}g^2(\frac{t_1}{2})^2 \rightarrow v_1 = \frac{gt_1}{2}
\end{equation}

Thus, $e$ can be rewritten in terms of the times between bounces:
\begin{equation}
    e = \frac{v_2}{v_1} = \frac{t_2}{t_1}
\end{equation}
Because we wanted to infer the parameter $e$, we chose $X_t = \frac{t_2}{t_1}$

In reality, perturbations occur at each bounce due to factors such as surface asperities. Consequently, the velocity of the ball has both horizontal and vertical components. Letting $d_1$ and $d_2$ be defined by the horizontal distances between bounces 1 and 2 and bounces 2 and 3, respectively, then $e$ would actually be derived from the ratio in Equation \ref{eq:2} as:

\begin{equation}
    e = \frac{v_2}{v_1} = \frac{\sqrt{v_{2_\text{horizontal}}^2 + v_{2_\text{vertical}}^2  }}{\sqrt{v_{1_\text{horizontal}}^2 + v_{1_\text{vertical}}^2  }}= \sqrt{\frac{(\nicefrac{d_2}{t_2})^2 + g^2(\nicefrac{t_2}{2})^2}{(\nicefrac{d_1}{t_1})^2 + g^2(\nicefrac{t_1}{2})^2}}
\end{equation}

However, since we capture positional information in the last three features of $X$, we decided to separate the temporal information from the positional information. Thus, we chose the simpler ratio of $e$ for $X_t$, which only depended on $t_1$ and $t_2$.

\subsection{Extracting $X_p$ from ball bounce locations}

The last three features of $X$, or $X_p$, encode positional information regarding the first three bounces. Letting $\vec{p}_1$ be the positional vector from the location of bounce 1 to the location of bounce 2, and $\vec{p}_2$ be the positional vector from the location of bounce 2 to the location of bounce 3 (see Figure \ref{fig:appendixa}), then the distances between the bounces can be calculated as:
\begin{equation}
    d_1 = ||\vec{p}_1||_2 \text{ and } d_2 = ||\vec{p}_2||_2
\end{equation}

We use $d_1$ and $d_2$ as the second and third features of $X$. Combined with the temporal information in $X_t$, a more accurate estimation of $e$ can be inferred. 

The last feature of $X_p$ encodes directional perturbations upon collision. Specifically, we find the signed angle $\alpha$ between $\vec{p}_1$ and $\vec{p}_2$ via the following expression:

\begin{equation}
    \alpha = \text{sign}(\frac{\vec{p}_{1_x} \vec{p}_{2_y} - \vec{p}_{2_x}\vec{p}_{1_y}}{||\vec{p}_1|| ||\vec{p}_2||})*\arccos(\frac{\vec{p}_1}{||\vec{p}_1||} \cdot \frac{\vec{p}_2}{||\vec{p}_2||})
\end{equation}

\section{Sound localization}
\label{app:soundlocalization}
In this section, we derive the Time Delay Estimation equations for 2D-localization using 3 microphones and compare the accuracy of offline and real-time sound localization methods. 

\subsection{Time Delay Estimation with 3 Microphones}

\begin{figure}[h]
    \centering
    \includegraphics[width=\textwidth]{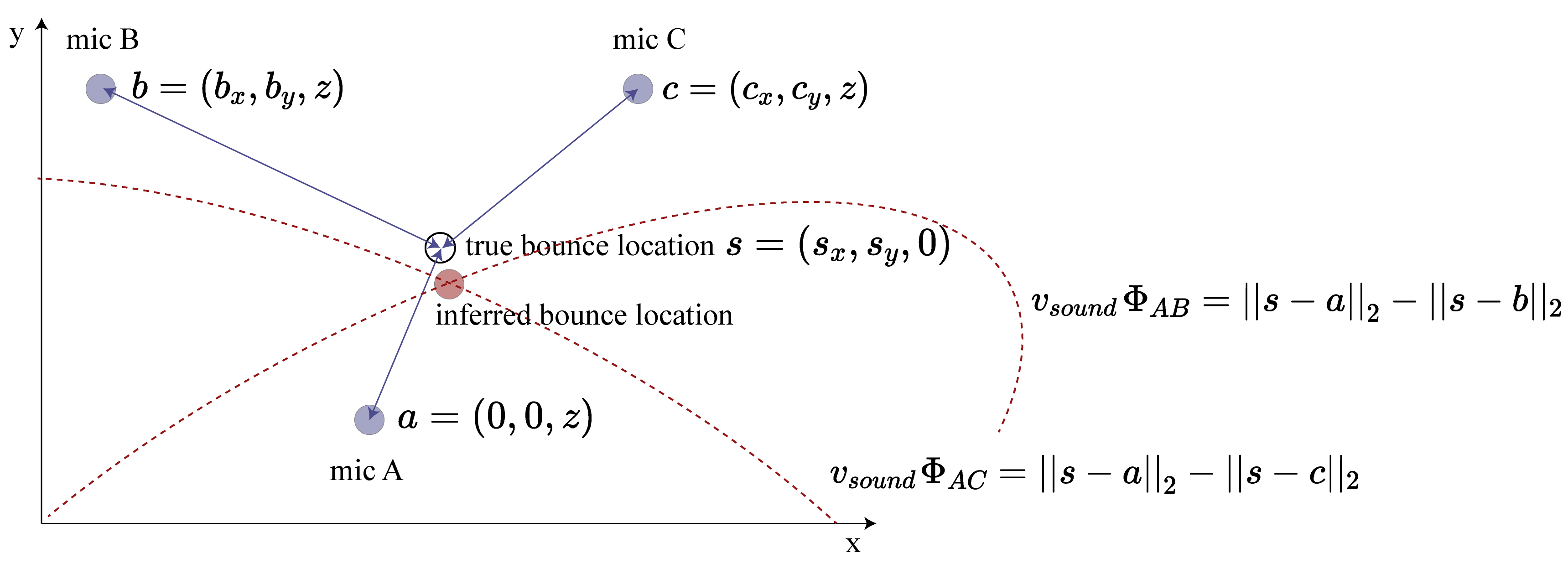}
    \caption{Locating the source of a sound using time delay estimation involves finding the intersection of two conics, defined by the time delays between pairs of microphones.}
    \label{fig:appendixb1}
    \end{figure}

Let $v_{sound}$ be the speed of sound through air, $\Phi_{AB}$ and $\Phi_{AC}$ be the time delays between signals received at microphones A and B and A and C, and denote the positions of the ball strike and microphones as $s=(s_x, s_y, 0)$, $a=(0,0,z)$, $b=(b_x, b_y, z)$, and $c=(c_x, c_y, z)$, respectively. Then the time delay $\Phi_{AB}$ is a function of the relative distances of the microphones to the ball strike, or

\begin{equation}
    \Phi_{AB}=\frac{1}{v_{sound}}(||s-a||_2 - ||s-b||_2)
\end{equation}

Thus, solving for the location of the ball strike is equivalent to finding the intersection of two conics (see Figure \ref{fig:appendixb1}):

\begin{equation}
    v_{sound}\Phi_{AB} = ||s-a||_2 - ||s-b||_2 \text{ and } v_{sound}\Phi_{AC} = ||s-a||_2 - ||s-c||_2
\end{equation}

A nonlinear optimizer is used to find this intersection. Because there can be more than one solution, we initialize the optimizer with a location inside the convex hull of the microphone positions.

\subsection{Comparison of Offline and Real-Time Sound Localization}

\begin{figure}[h]
    \centering
    \includegraphics[width=\textwidth]{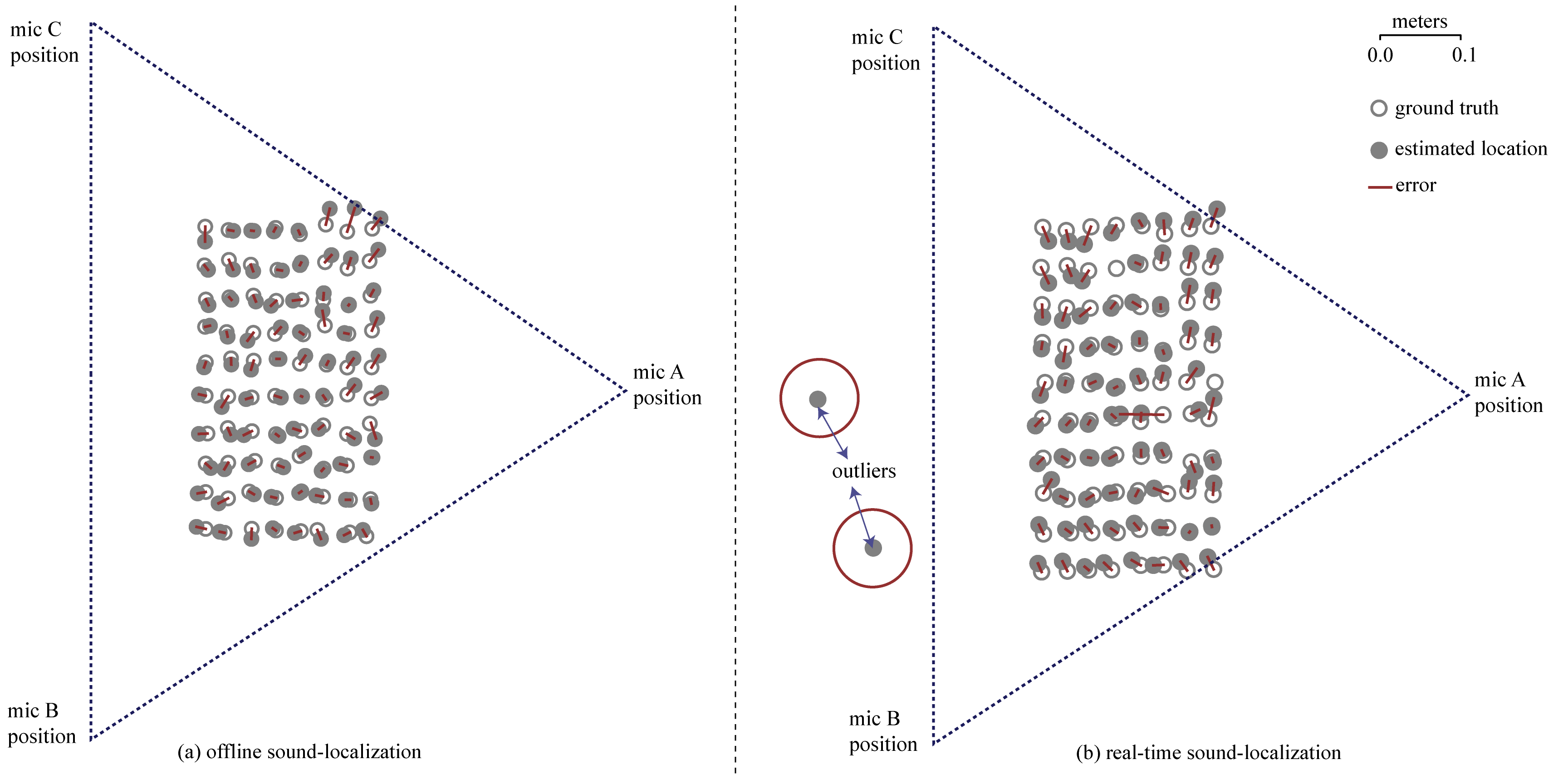}
    \caption{A comparison of (a) offline and (b) online sound localization. Ground truth positions were recorded by tapping a piece of paper layered with a carbon sheet. These positions were converted to world coordinates via an automated image processing script. Observe that outliers occasionally arise when using the less-accurate real-time sound localization method. However, the displacement errors of the outliers are sufficiently large such that they can be easily filtered.}
    \label{fig:appendixb2}
    \end{figure}

\section{Dynamic Prediction without Collision Noise}
\label{app:noiseless}

Below, we supplement the description of dynamic predictions without collision noise in Section \ref{sec:dynamic_predictions_nonoise} with relevant equations. The robot control sequence is summarized in Algorithm \ref{alg:1}.

\begin{figure}[h]
    \centering
    \includegraphics[width=0.36\textwidth]{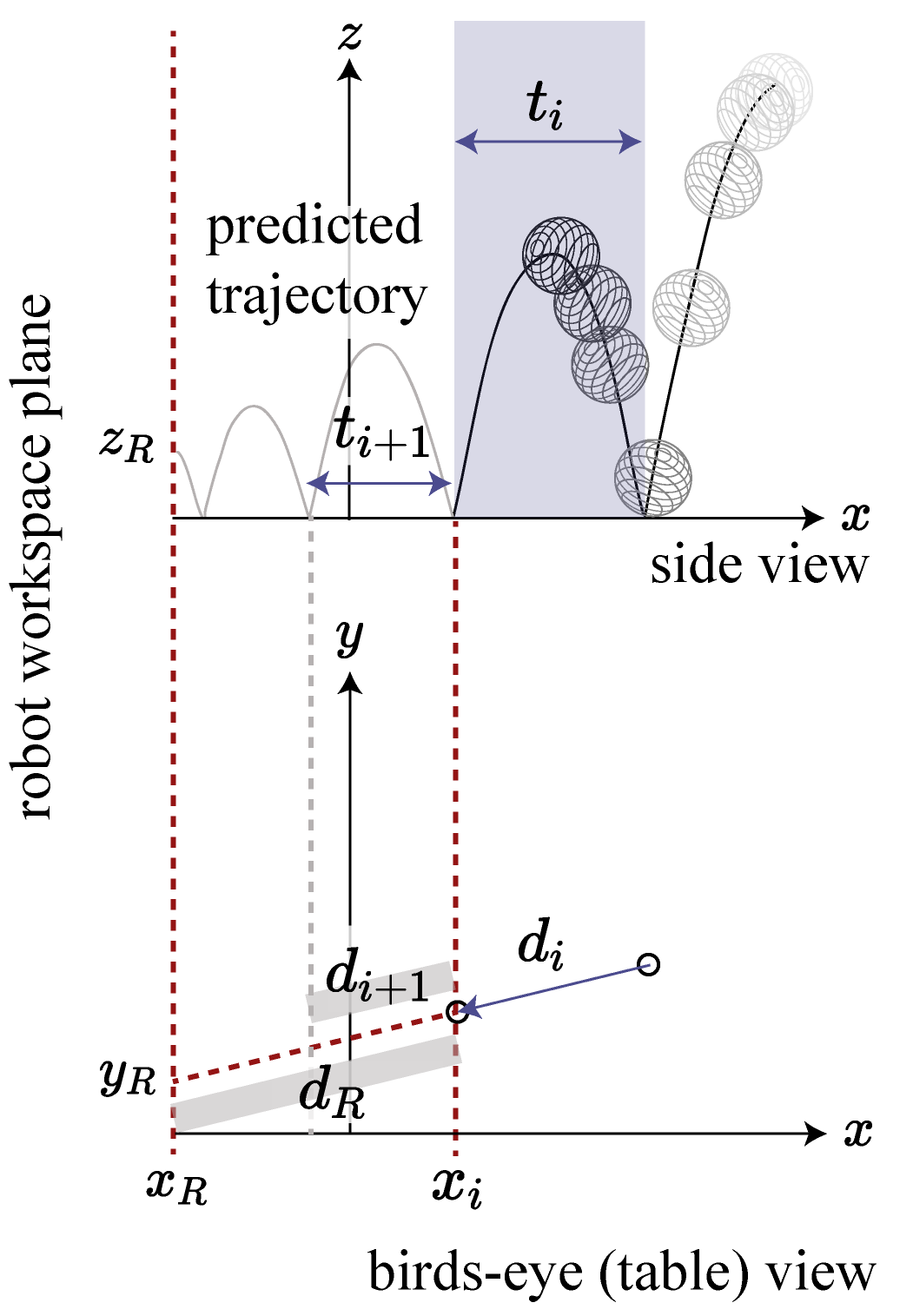}
    \caption{Dynamic predictions without collision noise assume a piecewise parabolic trajectory for the ball, based on the most recent two bounces. The robot moves to the intersection of this predicted trajectory and the robot workspace plane.}
    \label{fig:appendixc}
    \end{figure}

Let $(x_i, y_i)$, $t_i$, and $d_i$ be the position of the most recent bounce, time, and distance between the most recent and prior bounces (see Figure \ref{fig:appendixc}). Assuming that the ball travels in a piecewise parabolic trajectory with no directional perturbations upon each bounce, we can estimate the exit velocity of the ball rebounding from the bounce to be defined by: 
\begin{equation}
    v_{e(xy_i)}= \frac{ed_i}{t_i} \text{ and } v_{e(z_i)}= \frac{egt_i}{2}
\end{equation}
where $g$ is the gravitational constant, the $xy$ subscript refers to the horizontal velocity of the ball, and the $z$ subscript refers to the vertical velocity component of the ball.  These estimates allow us to calculate whether or not the ball will intersect the plane before the next bounce. First, denote $d_R$ as the horizontal distance from the most recent bounce to the robot plane along the direction of travel (see Figure \ref{fig:appendixc}). The distance and time of the next bounce relative to the most recent bounce, or $d_{i+1}$ and $t_{i+1}$, can be calculated by:
\begin{equation}
     d_{i+1} = e^2 d_i \text{ and } t_{i+1} = et_i
\end{equation}
Then to calculate whether or not the ball will intersect the robot plane before the next bounce, we check if
\begin{equation}
    d_{i+1} > d_R
\end{equation}
 If true, the ball will intersect the plane before it lands again in $t_R$ seconds, so the end-effector must be moved to the intersection point $(x_R, y_R, z_R)$ in $t_R$ seconds. If the ball will not intersect the plane before it lands again, the end-effector moves to $y_R$ in $t_{i+1}$ seconds before the next bounce is observed. 
 
 The intersection point $(x_R, y_R, z_R)$ is partially pre-determined and therefore fixed by the placement of the robot (which sits at $x=x_R$). We can calculate $y_R$ by finding the intersection between line $x=x_R$ and the line connecting the last two ball bounce positions (see Figure \ref{fig:appendixc}). This intersection occurs at: 

\begin{equation}
    y_R = \frac{y_i - y_{i-1}}{x_i - x_{i-1}}x_R + (y_i - \frac{y_i - y_{i-1}}{x_i - x_{i-1}}x_i)
\end{equation}

Finally, if it is determined that the ball will intersect the robot workspace plane before it bounces again, classical projectile equations can be used to calculate $t_R$ and $z_R$:

\begin{equation}
    t_R = \frac{d_R}{v_{e_{xy_i}}} = \frac{d_R t_i}{ed_i} 
\end{equation}

\begin{equation}
    z_R =  v_{e(z_i)} t_R - \frac{1}{2}gt_R^2 = (\frac{egt_i}{2})(\frac{d_R t_i}{ed_i}) - \frac{1}{2}g(\frac{d_R t_i}{ed_i})^2 
\end{equation}

\vspace{55pt}
\begin{algorithm}[H]
\SetAlgoLined
\KwResult{Positions the end-effector of a robotic arm to meet a bouncing ball}
 \While{Received bounce position and times}{
 $(x,y) \leftarrow$ bounce location\;
  \eIf{First bounce }{
  Move end-effector to position $y$\;
  }{
  \eIf{After next bounce, the ball will intersect the robot plane at $(y_R, z_R)$}{
  Move end-effector to position $(y_R, z_R)$ in $t_R$ seconds\;}{
  Move end-effector to $y_R$ in $t_{i+1}$ seconds (before the next bounce is observed)\;}
  }
 }
 \caption{Robot Control using Dynamic Predictions without Collision Noise}
 \label{alg:1}
\end{algorithm}

\section{Dynamic Predictions with Collision Noise}
\label{app:noisy}

In this section, the algorithm for making dynamic predictions with collision noise is outlined in Algorithm \ref{alg:2}. Additionally, a complete breakdown of all 30 tosses per ball is included, and the failure modes due to outliers are compared between Algorithm \ref{alg:1} and Algorithm \ref{alg:2}.  Finally, two real examples are shown with freeze-frames in Section \ref{sec:freezeframes} to detail the algorithmic steps associated with each bounce event. 

\begin{algorithm}[H]
\SetAlgoLined
\KwResult{Positions the end-effector of a robotic arm to meet a bouncing ball}
 \While{Received bounce position and times}{
  $(x,y) \leftarrow$ bounce location\;
  \uIf{First bounce}{
     Move end-effector to position $y$\;
  }
  \uElseIf{Second bounce}{
    next\_bounce\_posterior $\leftarrow$ Generate posterior for next bounce position and time \;
    Look ahead $k$ bounces to generate $n$ samples where the ball will intersect the plane \;
    \uIf{Any samples intersect the plane of the robot}{Move $\propto \sigma^{-1}$ distance toward mean of samples, where $\sigma$ is the standard deviation of the intersection points on the robot plane \;}
  }
  \Else{
    \uIf{Bounce position is an outlier}{Replace bounce location with the mean of next\_bounce\_posterior \;}
    Repeat lines 6-9 \;
    
  }
 }
 \caption{Robot Control using Dynamic Predictions with Collision Noise}
 \label{alg:2}
\end{algorithm}

As discussed in Section \ref{sec:dynamicpredictions_collisionnoise}, a new BayesSim model is trained to generate posteriors over the next bounce time and location, given the time and distance between the prior two bounces. After the second bounce is observed, this new BayesSim model can be used to generate a posterior for the next bounce position and time. Furthermore, the model can be used to recursively sample future bounce positions until they pass the robot workspace plane. For example, let $k$ be the number of bounces to look ahead and $n$ be the number of samples to generate at each future bounce. Given the last two bounces $i$ and $i-1$, a posterior over the future bounce $i+1$ is generated using the BayesSim model. This posterior is sampled $n$ times, and each sample can be used to create a new posterior with the most recent bounce $i$. Thus, $n$ posteriors over bounce $i+1$ are generated. This is repeated up to $k$ times until the sample intersects the workspace plane. A maximum of $n^k$ samples are generated in this step. In experiments, we found that $n=10$ and $k=[2,4]$ were sufficient at capturing the uncertainty in future bounce locations and did not significantly add to the latency of the entire algorithm (this inference step ran at about 20 Hz). We varied $k$ depending on how reactive we wanted the robot to be. For instance, the moonball is highly uncertain at each collision, so we used $k=2$. In contrast, since the ping-pong ball experienced lower collision noise, we used $k=4$.

At every subsequent bounce after the second bounce, the bounce location is classified as either an outlier or an inlier by using an elliptic envelope filter with the future bounce posterior generated at the prior bounce. If the bounce location is determined to be an outlier, then the mean of that posterior is used in place of the outlier location. This extra check was implemented to compensate for inaccurate real-time bounce localization due to high reverberation. In the following section, we compare the sensitivity to outliers using Algorithms \ref{alg:1} and \ref{alg:2}.

\subsection{Full Breakdown of Success and Failures out of all 30 Tosses}
\label{sec:appendixd1}

\begin{figure}[h]
    \centering
    \includegraphics[width=0.9\textwidth]{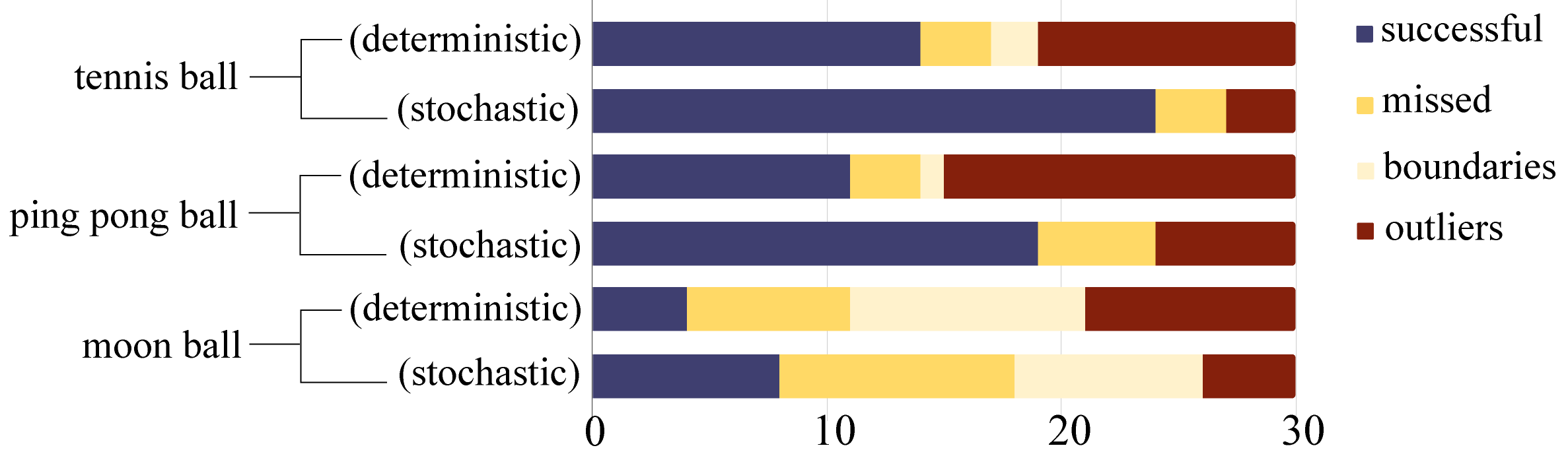}
    \caption{"Outliers" signifies failures due to errors in sound localization, "Boundaries" refers to failures due to the ball violating robot workspace boundaries, and "Missed" denotes any other failure. All 30 tosses are represented in this graph.}
    \label{fig:appendixd1}
    \end{figure}
    
Section \ref{sec:roboticballtrackingresults} performs analysis on the tosses that do not violate the robot's workspace constraints. For completeness, we include a full breakdown of all 30 tosses in Figure \ref{fig:appendixd1}, which includes the failures due to the ball violating the robot workspace boundaries. As shown in Figure \ref{fig:appendixd1}, few to none of the tosses violate the boundary for the tennis ball and ping pong ball, but almost a third of the tosses for the moon ball result in boundary violations. This is because the moonball experiences highly stochastic collisions. 

\subsection{Comparison of Sensitivity to Outliers with Algorithms 1 and 2}

\begin{figure}[h]
    \centering
    \includegraphics[width=0.9\textwidth]{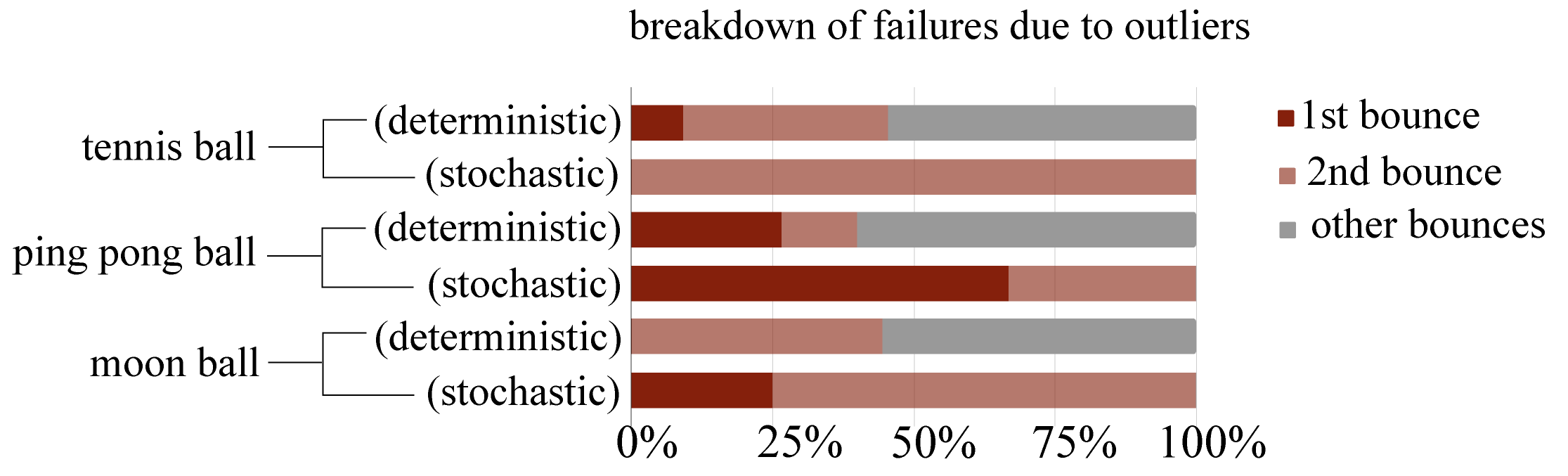}
    \caption{We categorize the failures due to outliers in bounce localization by the bounce at which the outlier occurs. Note that Algorithm 2 successfully filters out outliers as long as they occur after the second bounce. The robot is able to use, instead, the mean of the posterior over which the bounce \textit{should} have occurred, which subsequently leads to more accurate ball trajectory predictions.}
    \label{fig:appendixd2}
    \end{figure}

In Figure \ref{fig:appendixd2}, we break down the failure modes of both Algorithms \ref{alg:1} and \ref{alg:2} due to outliers in sound localization. As shown in the graph, both methods are sensitive to outliers in the first or second bounce. Because Algorithm \ref{alg:2} relies on accurate first and second bounce locations to predict future bounce distributions, there currently is not a way to adjust for these errors. We believe that we can eliminate these outliers by implementing more efficient real-time processing, for instance, by using a real-time audio-processing language called SuperCollider, or by training a neural network to perform fast peak detection. However, by considering collision noise for dynamic predictions, Figure \ref{fig:appendixd2} shows that Algorithm \ref{alg:2} successfully filters out outliers in any subsequent bounces and can compensate for their errors. 

\subsection{Example robot trials with captions}\label{sec:freezeframes}

Below, we detail two real robotic examples with freeze-frames to depict the steps of Algorithm \ref{alg:2}. 

\subsubsection{Example 1: Success with no outliers}

\begin{figure}[h]
    \centering
    \includegraphics[width=0.7\textwidth]{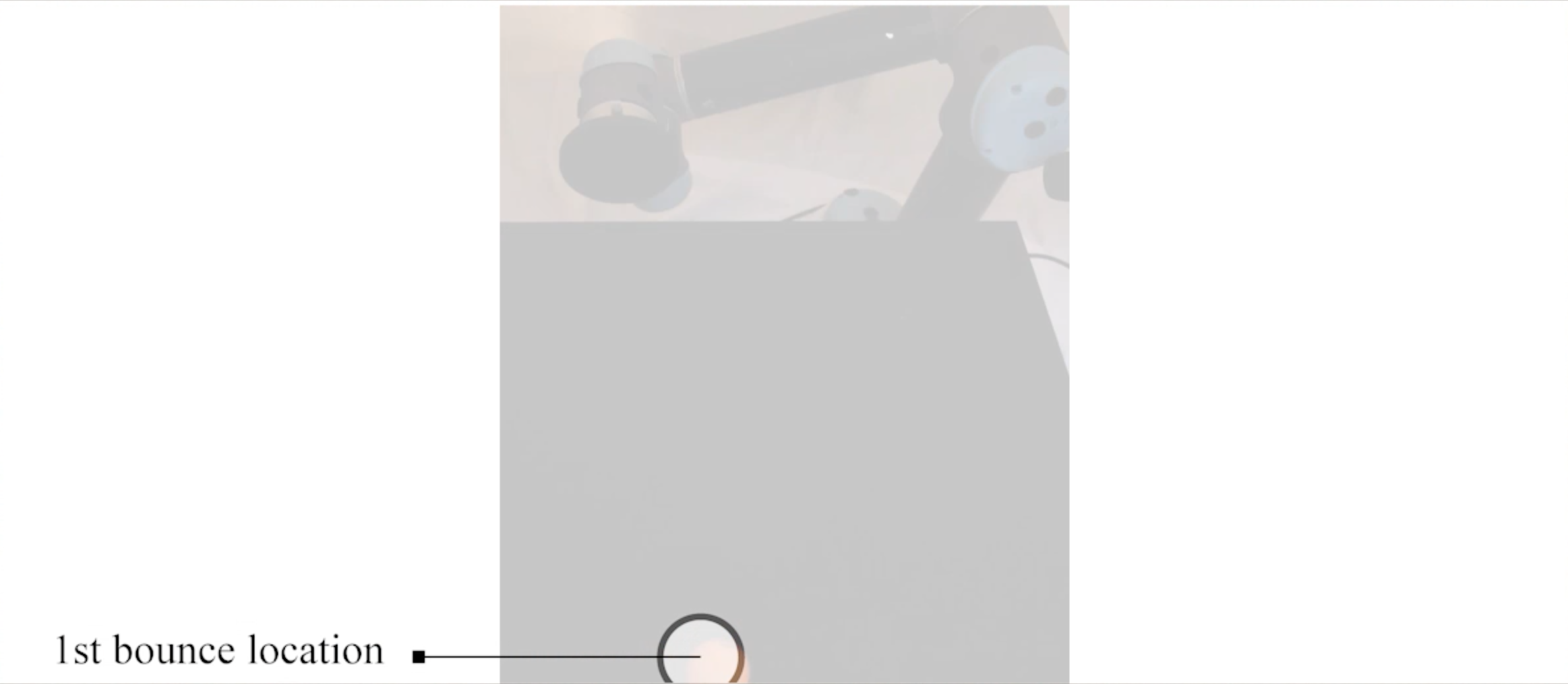}
    \caption{The first bounce is located.}
    \label{fig:23_1}
    \end{figure}
\begin{figure}[h]
    \centering
    \includegraphics[width=0.7\textwidth]{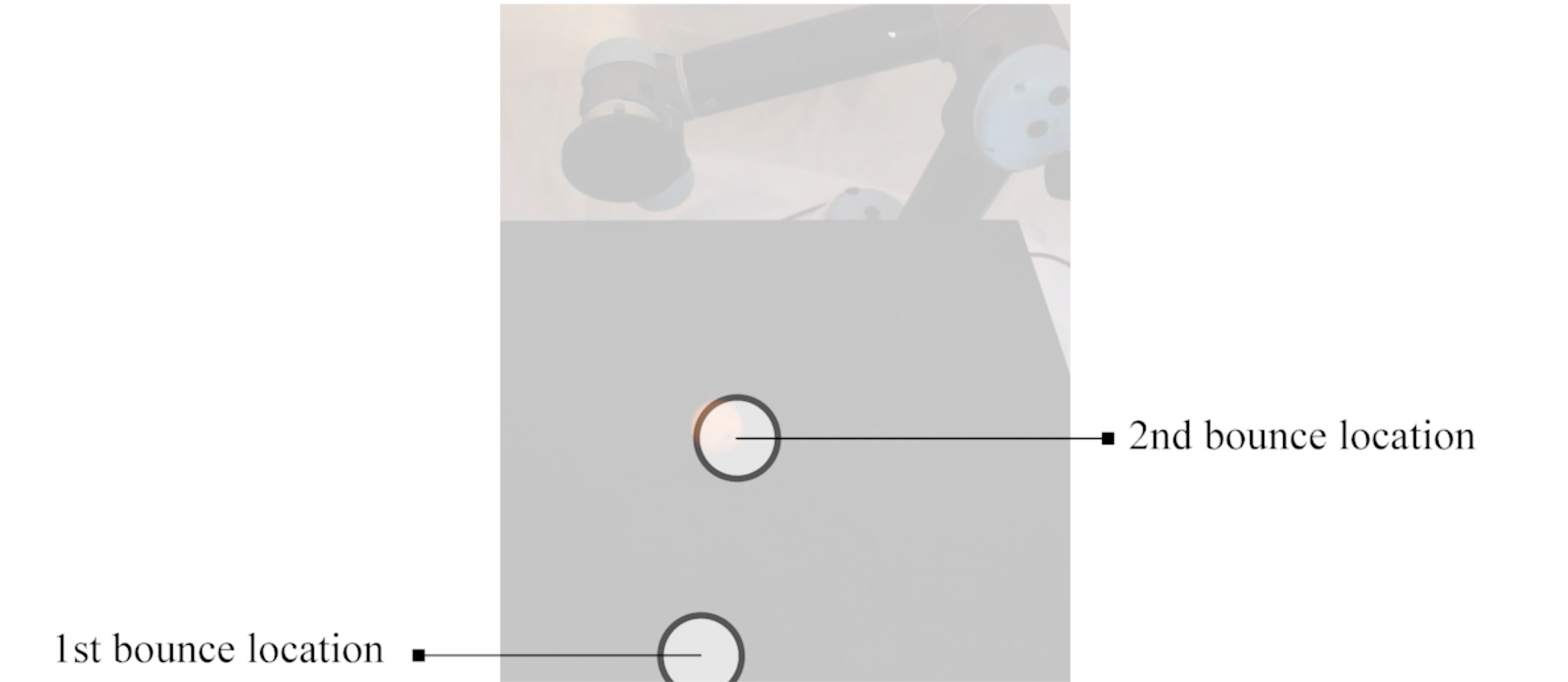}
    \caption{The second bounce is located.}
    \label{fig:23_2}
    \end{figure}
\begin{figure}[h]
    \centering
    \includegraphics[width=0.7\textwidth]{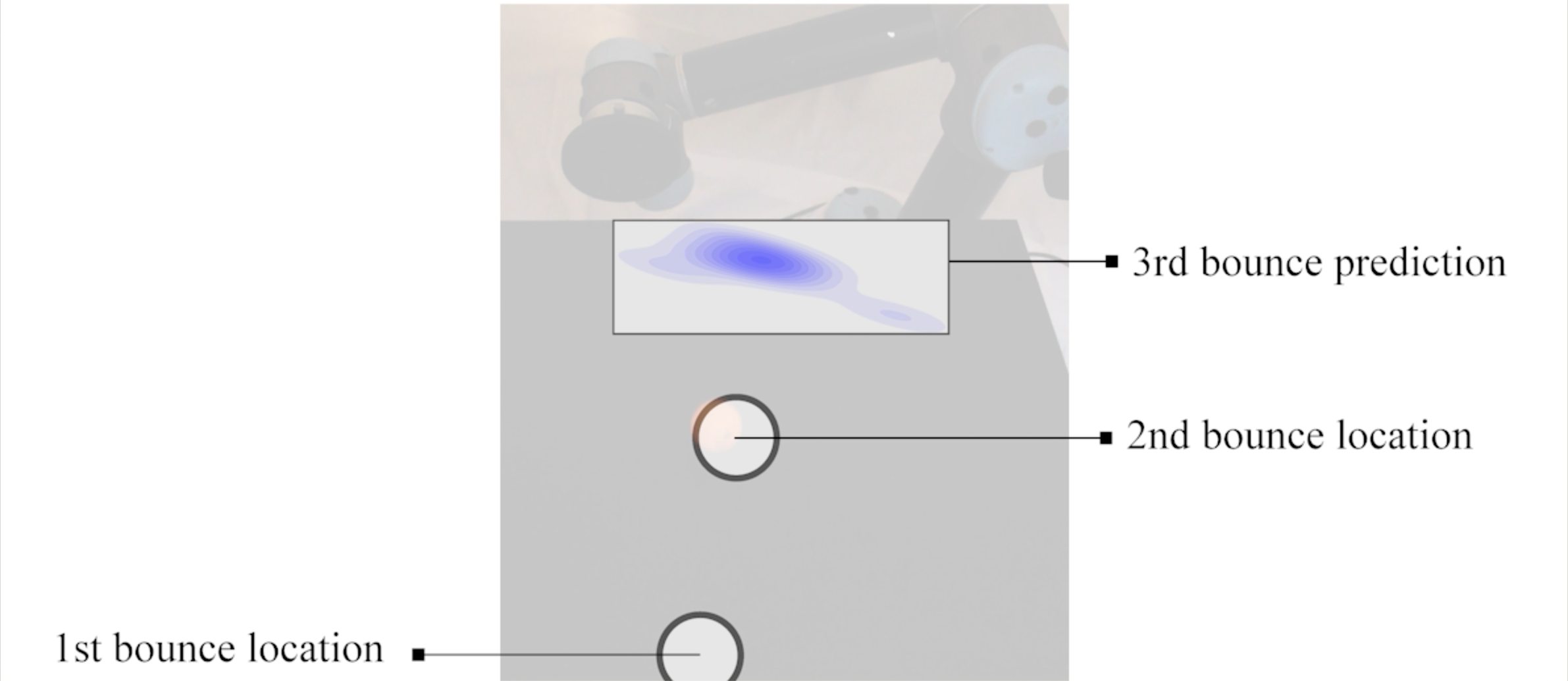}
    \caption{The 1st and 2nd measured locations are used to predict the distribution over the 3rd bounce location.}
    \label{fig:23_3}
    \end{figure}
\begin{figure}[h]
    \centering
    \includegraphics[width=0.7\textwidth]{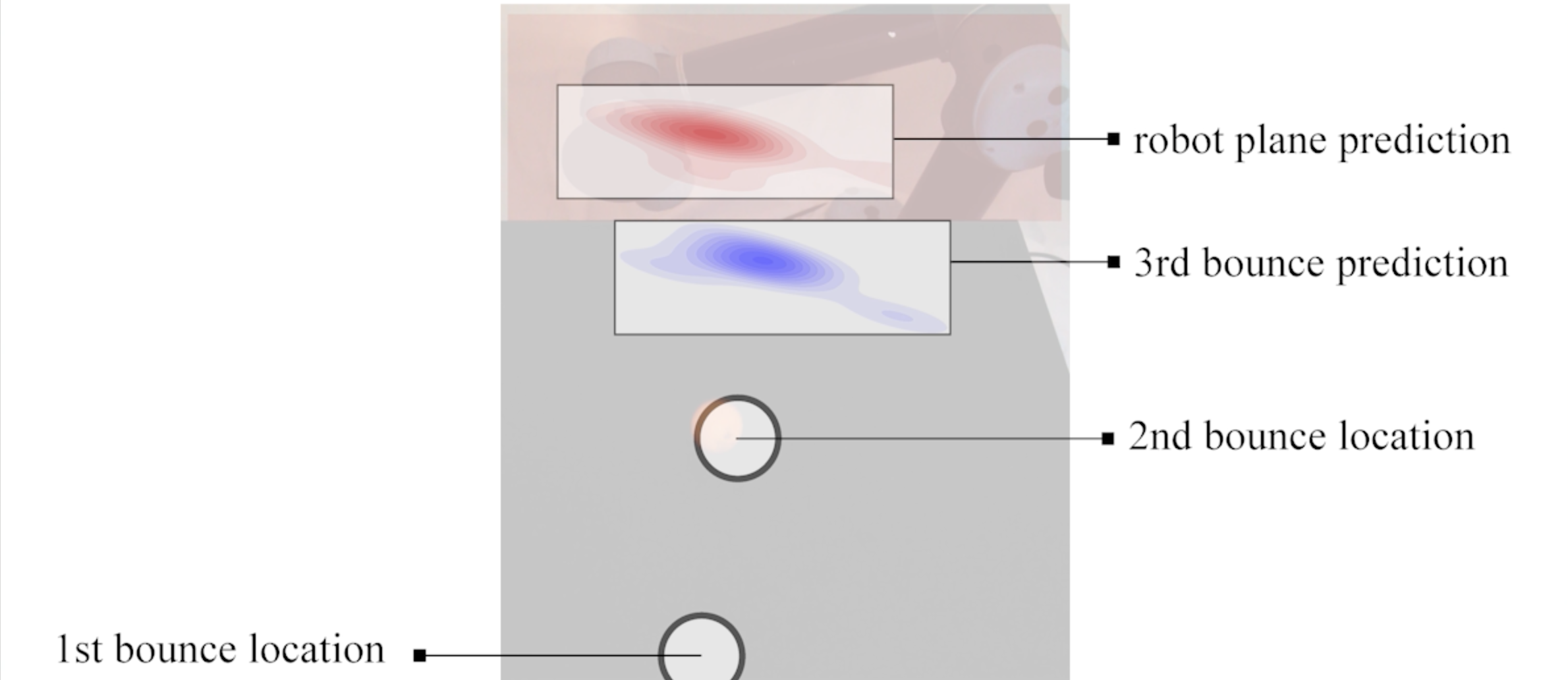}
    \caption{The 2nd measured location and the 3rd bounce location distribution are used to predict the distribution of the robot plane intersections.}
    \label{fig:23_4}
    \end{figure}
\begin{figure}[h]
    \centering
    \includegraphics[width=0.7\textwidth]{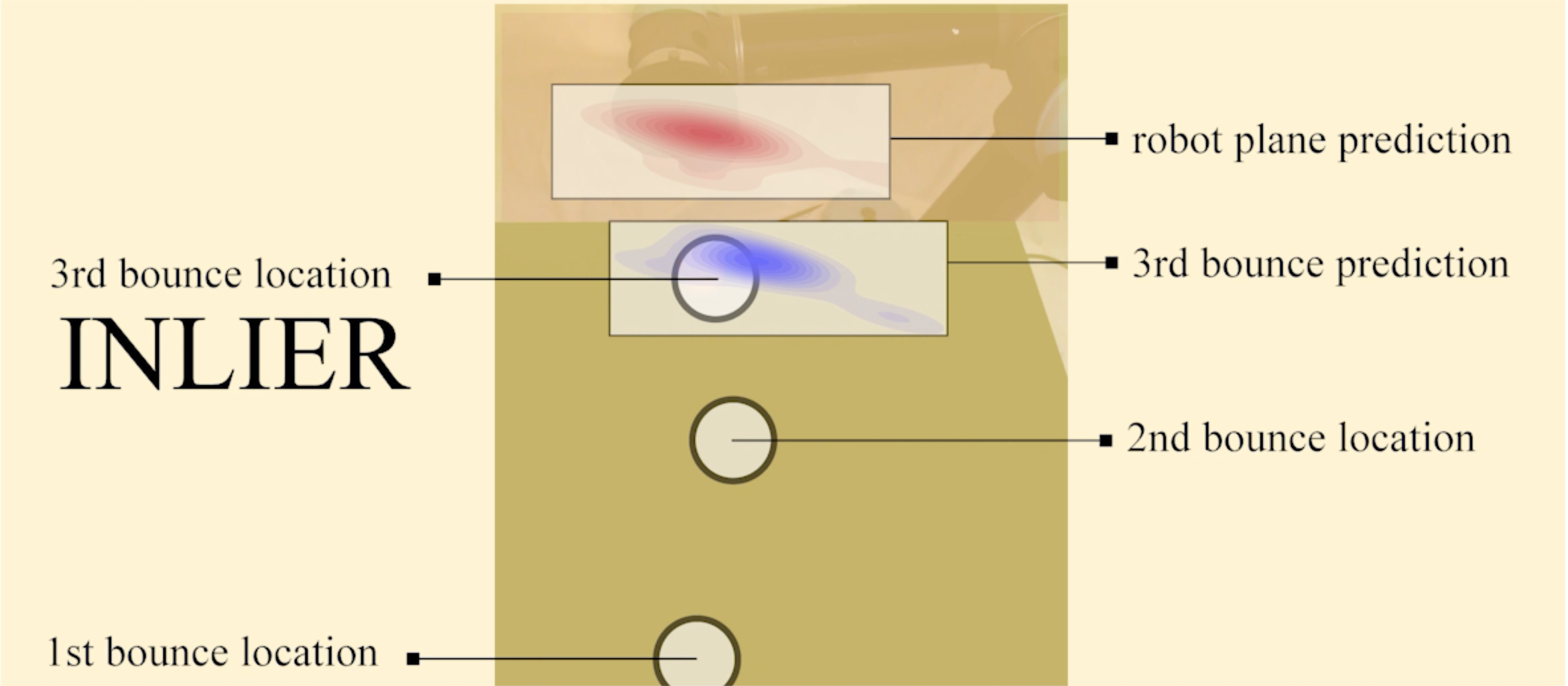}
    \caption{Inlier: the audio-based localization of the 3rd bounce lies within the predicted distribution and is trusted. This leads to a successful prediction of the robot plane intersection, and the robot comes into contact with the ball}
    \label{fig:23_5}
    \end{figure}
    
\clearpage

\subsubsection{Example 2: Success with an outlier}

\begin{figure}[h]
    \centering
    \includegraphics[width=0.7\textwidth]{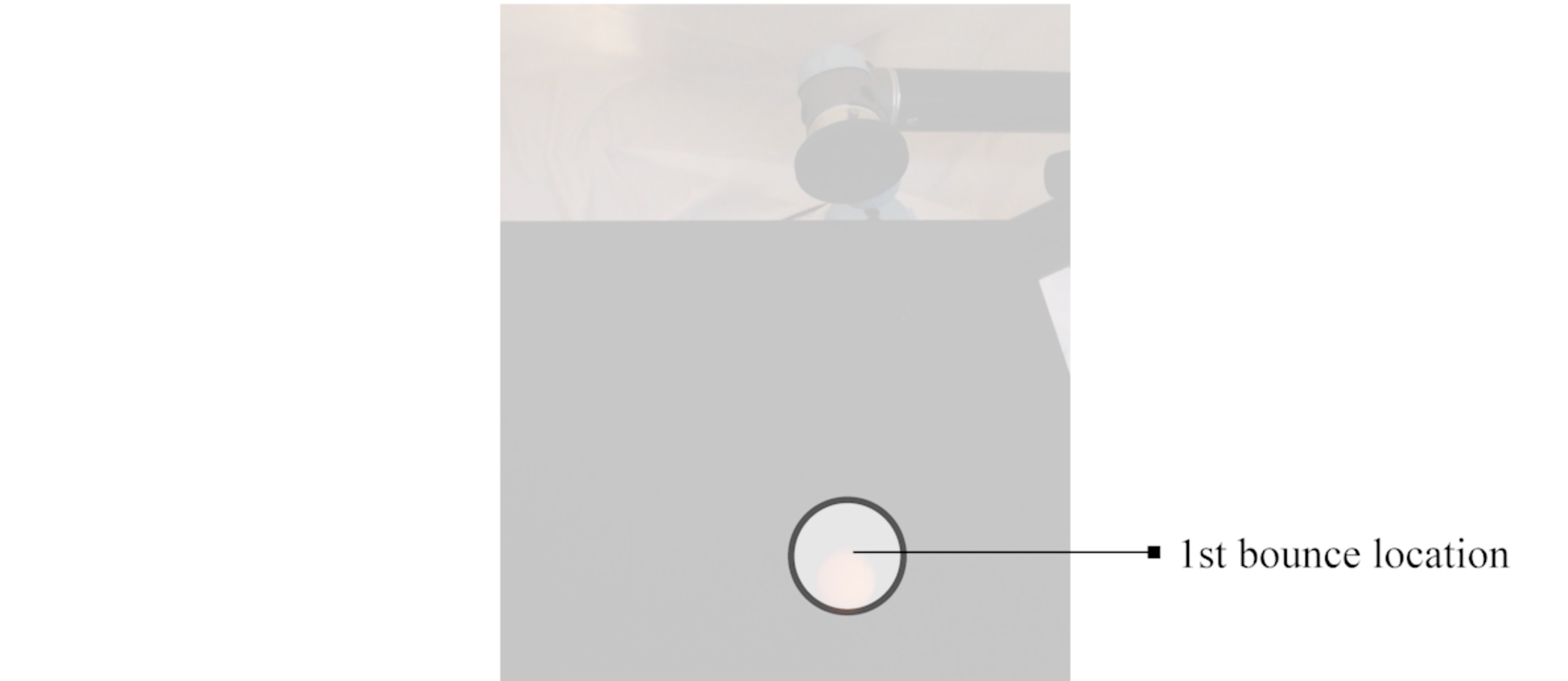}
    \caption{The first bounce is located.}
    \label{fig:21_1}
    \end{figure}
    
\begin{figure}[h]
    \centering
    \includegraphics[width=0.7\textwidth]{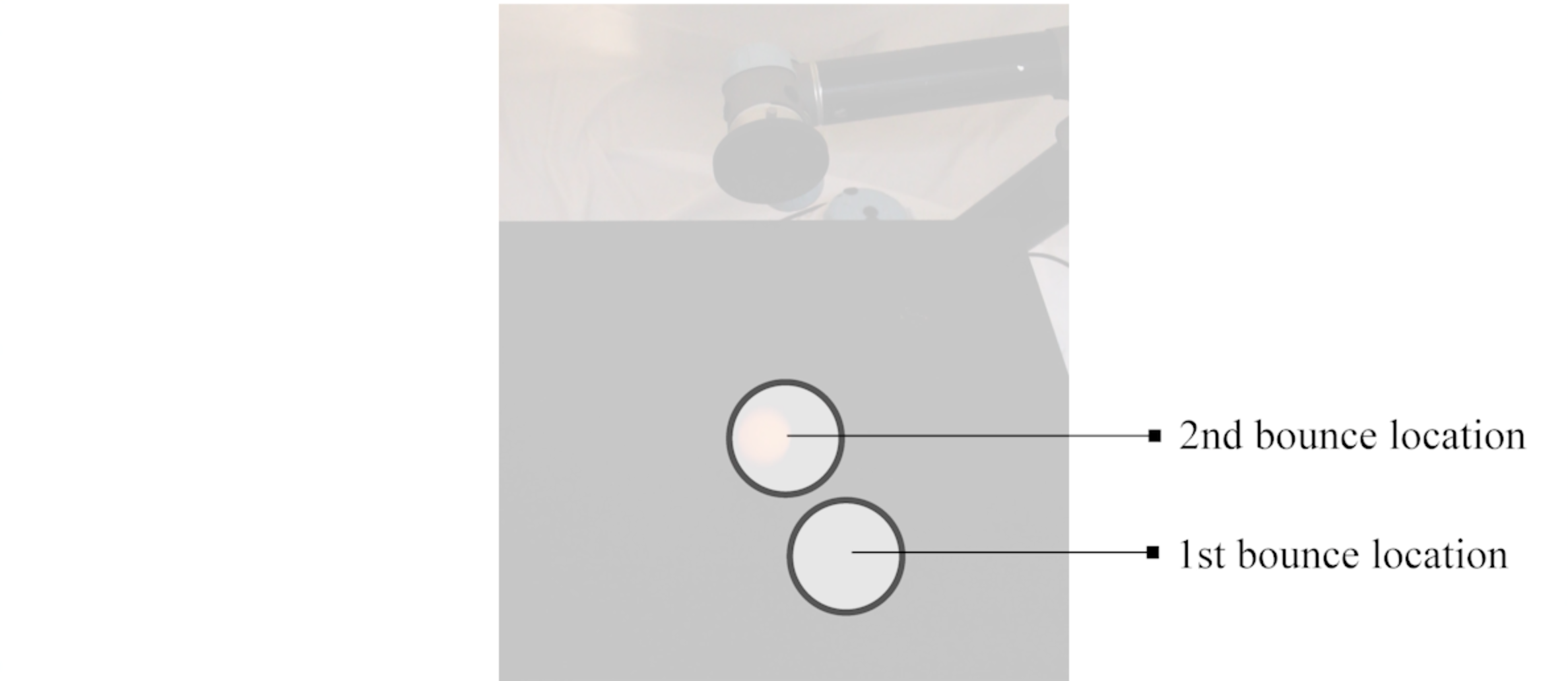}
    \caption{The second bounce is located.}
    \label{fig:21_2}
    \end{figure}
\begin{figure}[h]
    \centering
    \includegraphics[width=0.7\textwidth]{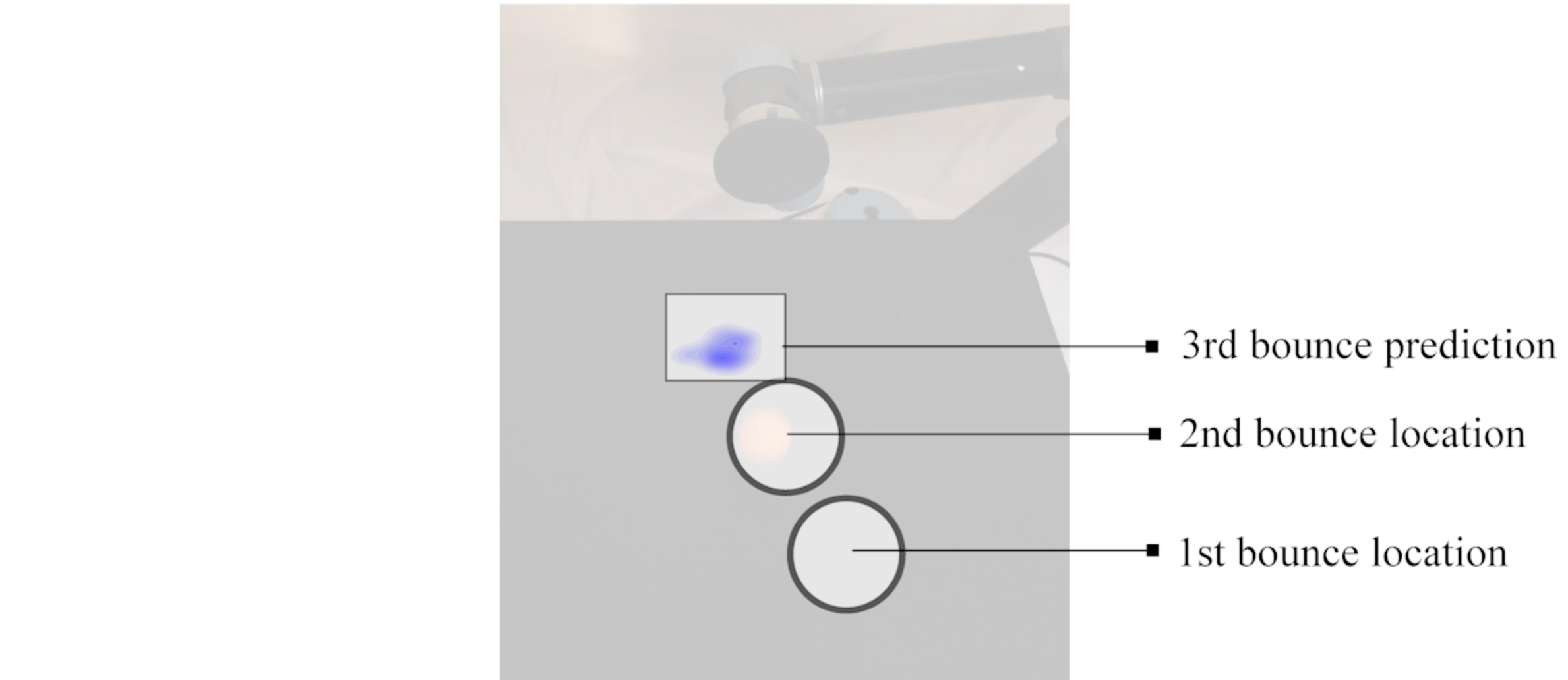}
    \caption{The 1st and 2nd measured locations are used to predict the distribution over the 3rd bounce location.}
    \label{fig:21_3}
    \end{figure}
\begin{figure}[h]
    \centering
    \includegraphics[width=0.7\textwidth]{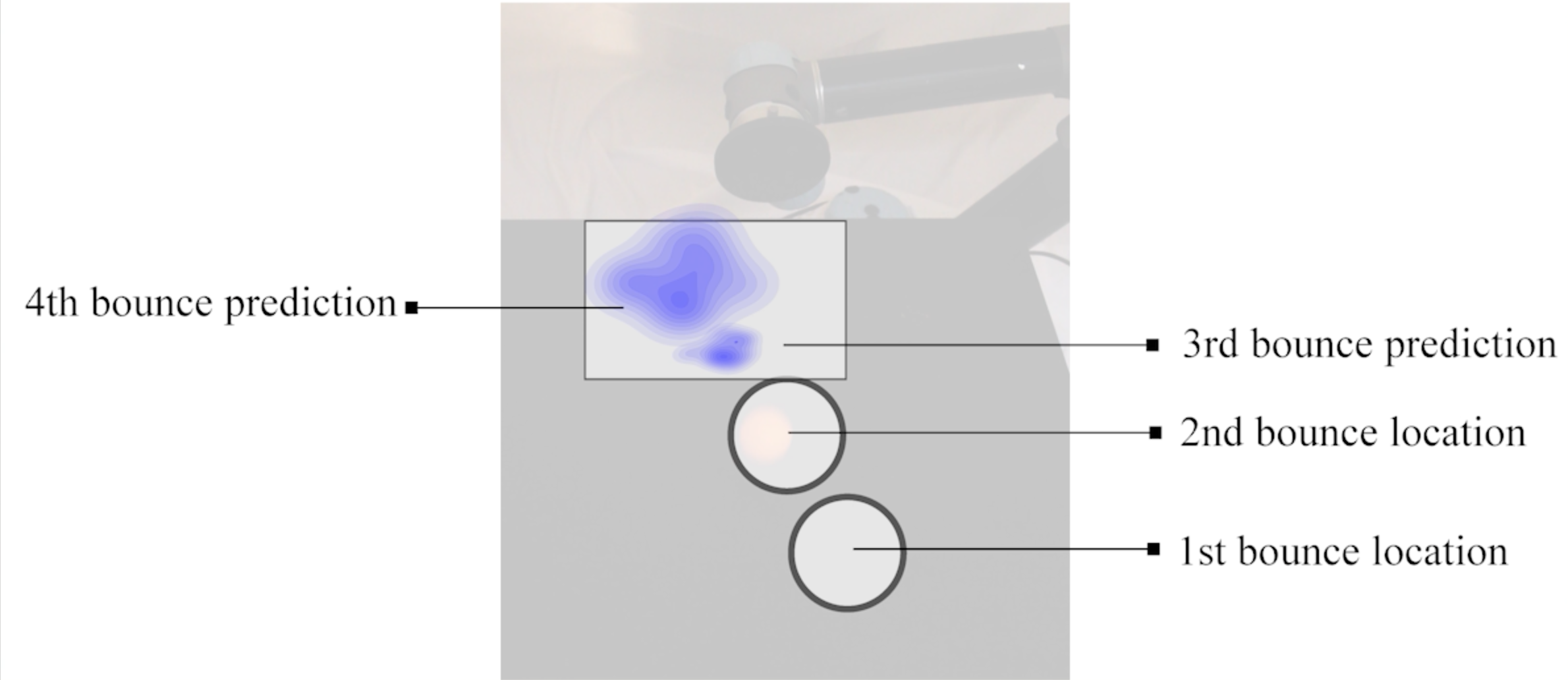}
    \caption{The 2nd measured location and 3rd bounce location distribution are used to predict the distribution over the 4th bounce location.}
    \label{fig:21_4}
    \end{figure}
    
\begin{figure}[h]
    \centering
    \includegraphics[width=0.7\textwidth]{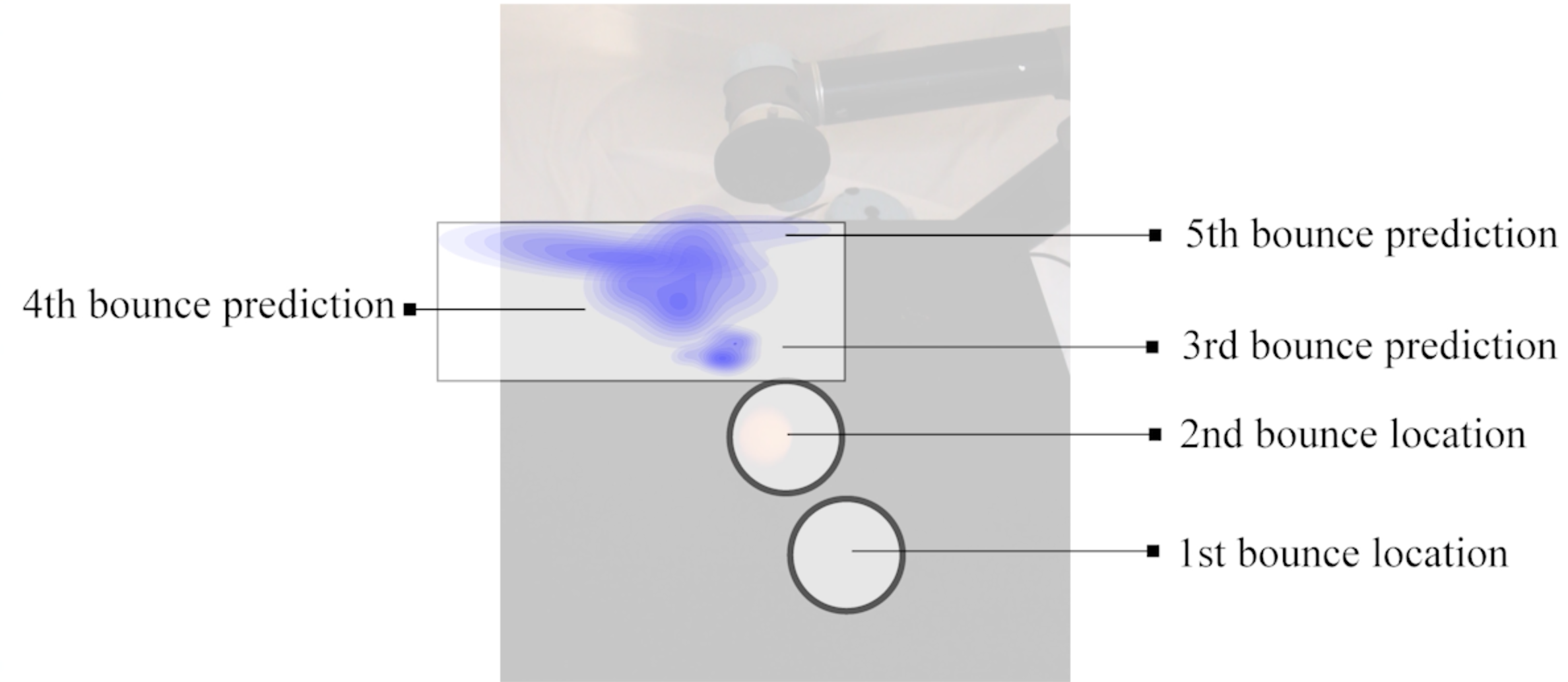}
    \caption{The 3rd and 4th bounce location distributions are used to predict the distribution over the 5th bounce location.}
    \label{fig:21_5}
    \end{figure}
    
\begin{figure}[h]
    \centering
    \includegraphics[width=0.7\textwidth]{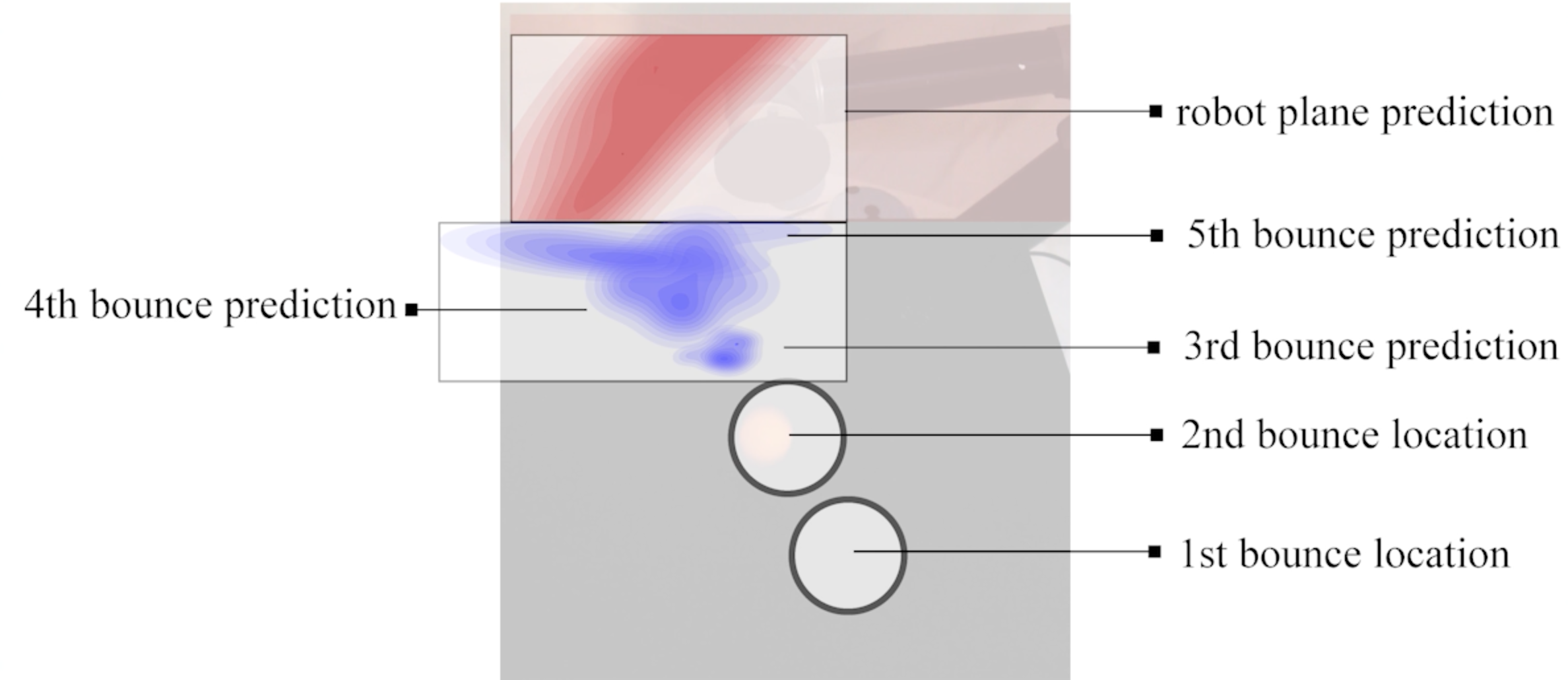}
    \caption{The 4th and 5th bounce location distributions are used to predict the distribution of the robot plane intersections.}
    \label{fig:21_6}
    \end{figure}
    
\begin{figure}[h]
    \centering
    \includegraphics[width=0.7\textwidth]{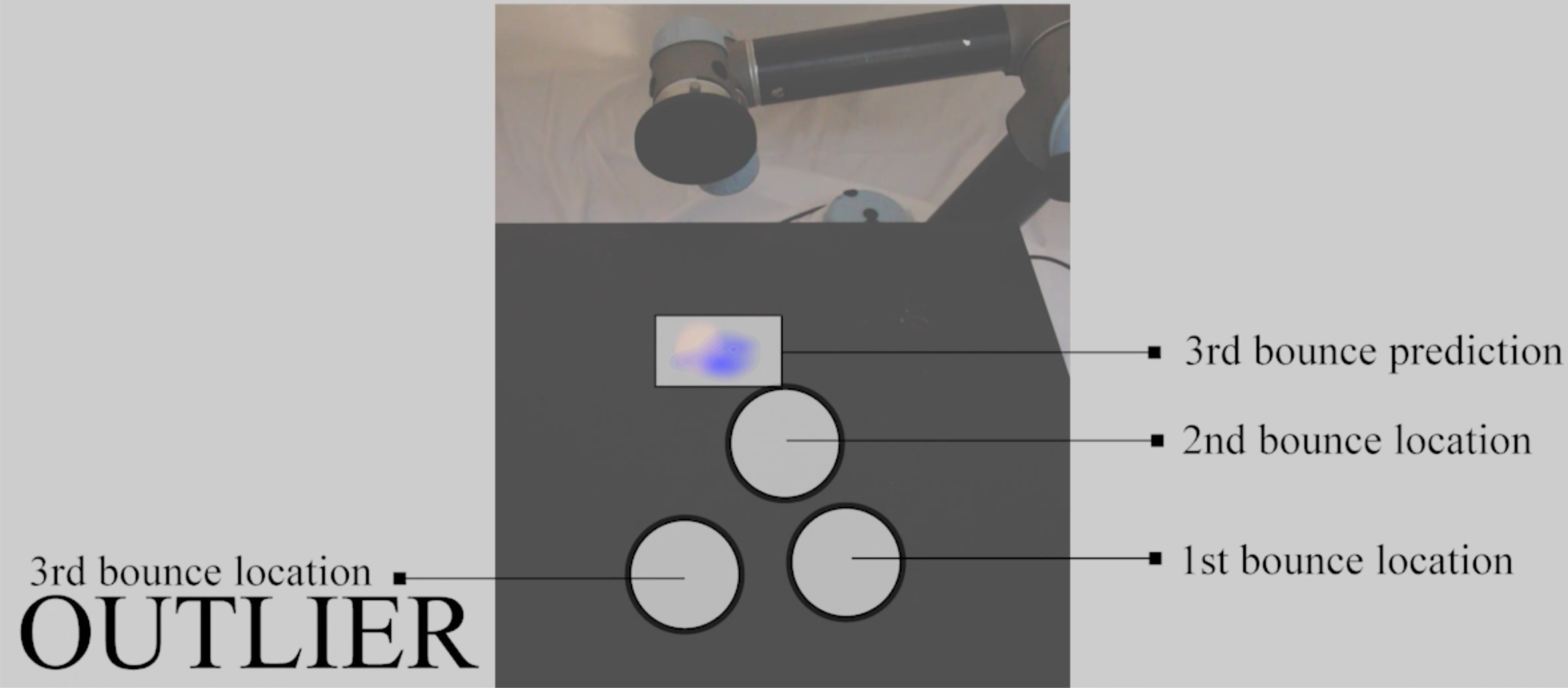}
    \caption{Outlier: the audio-based localization of the 3rd bounce lies far from the predicted distribution and is ignored.}
    \label{fig:21_7}
    \end{figure}
    
\begin{figure}[h]
    \centering
    \includegraphics[width=0.7\textwidth]{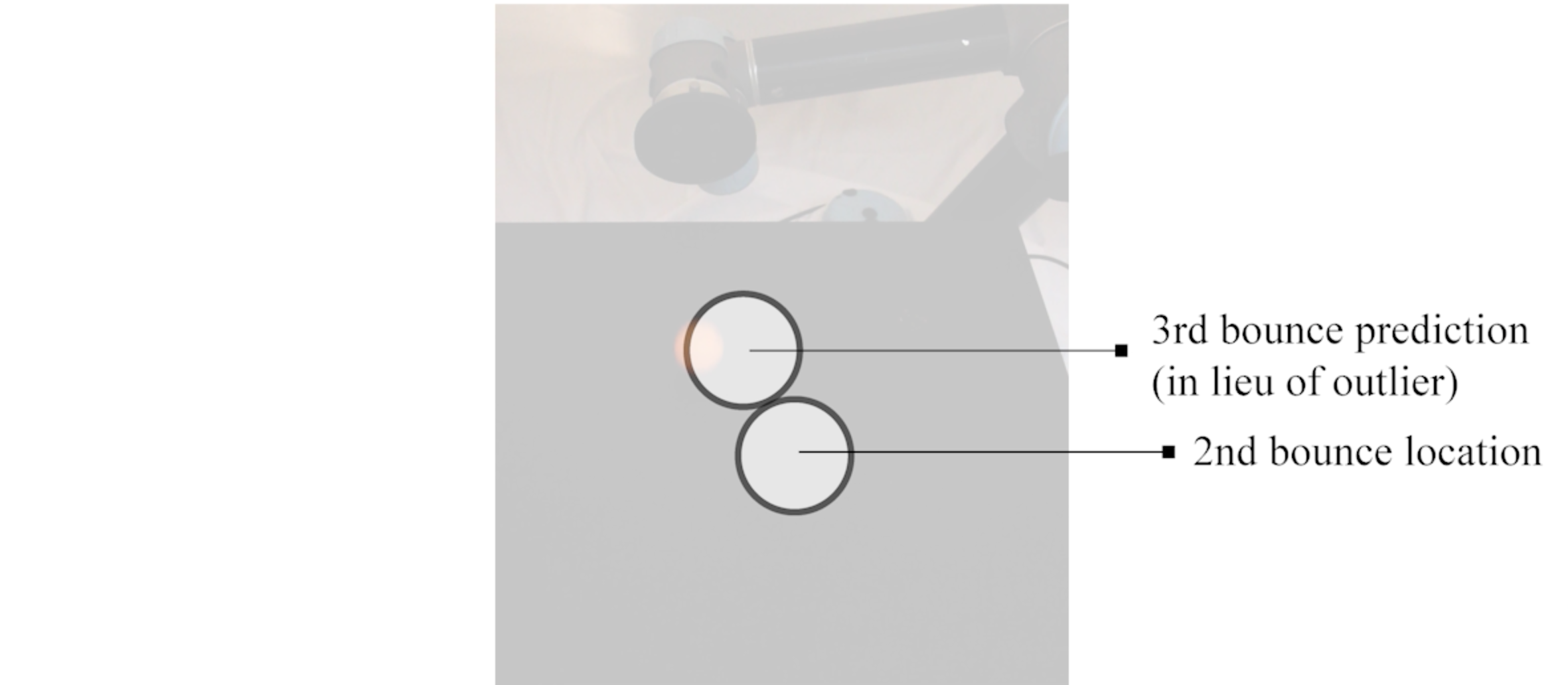}
    \caption{The mean of the predicted distribution for the 3rd bounce is used in lieu of the outlier.}
    \label{fig:21_8}
    \end{figure}
    
\begin{figure}[h]
    \centering
    \includegraphics[width=0.7\textwidth]{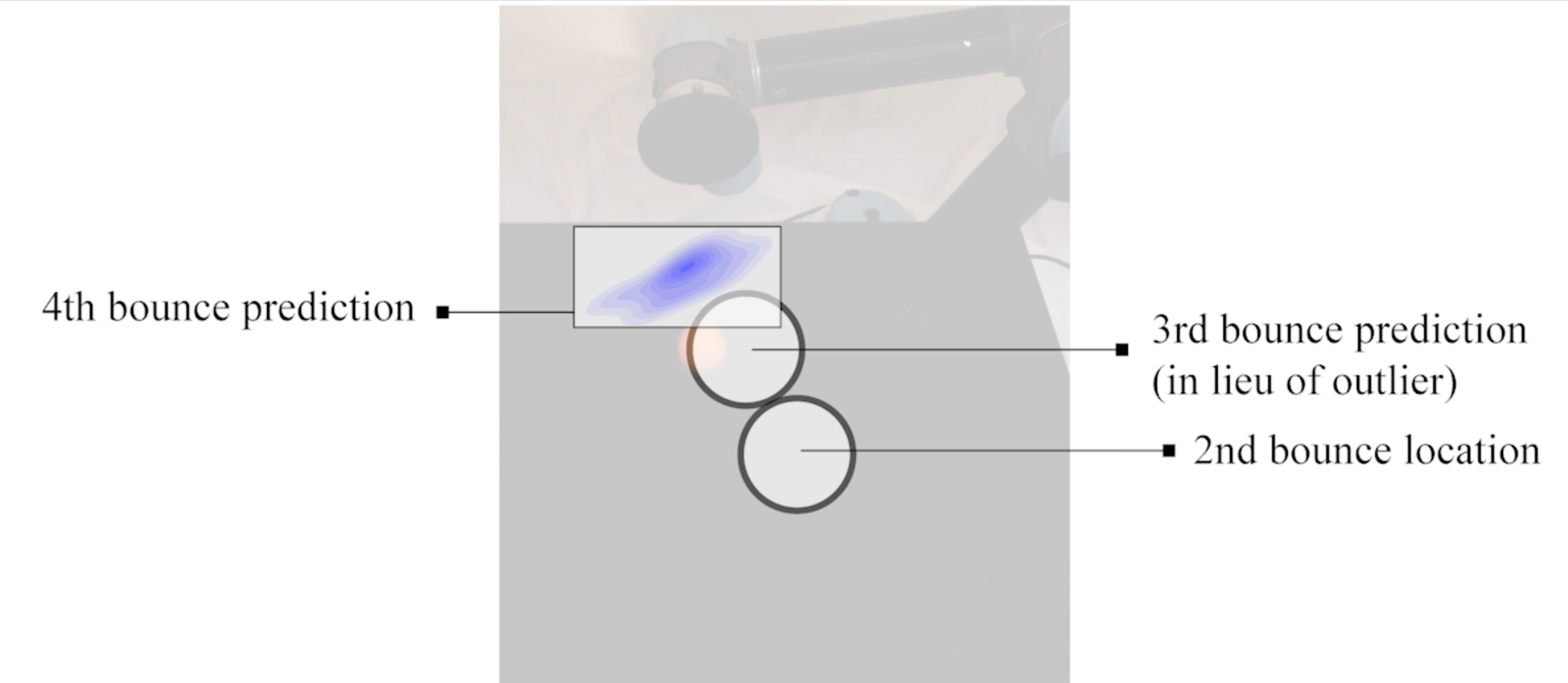}
    \caption{The 2nd measured location, as well as the 3rd \textit{predicted} location, are used to predict the distribution of the 4th bounce.}
    \label{fig:21_9}
    \end{figure}
    
\begin{figure}[h]
    \centering
    \includegraphics[width=0.7\textwidth]{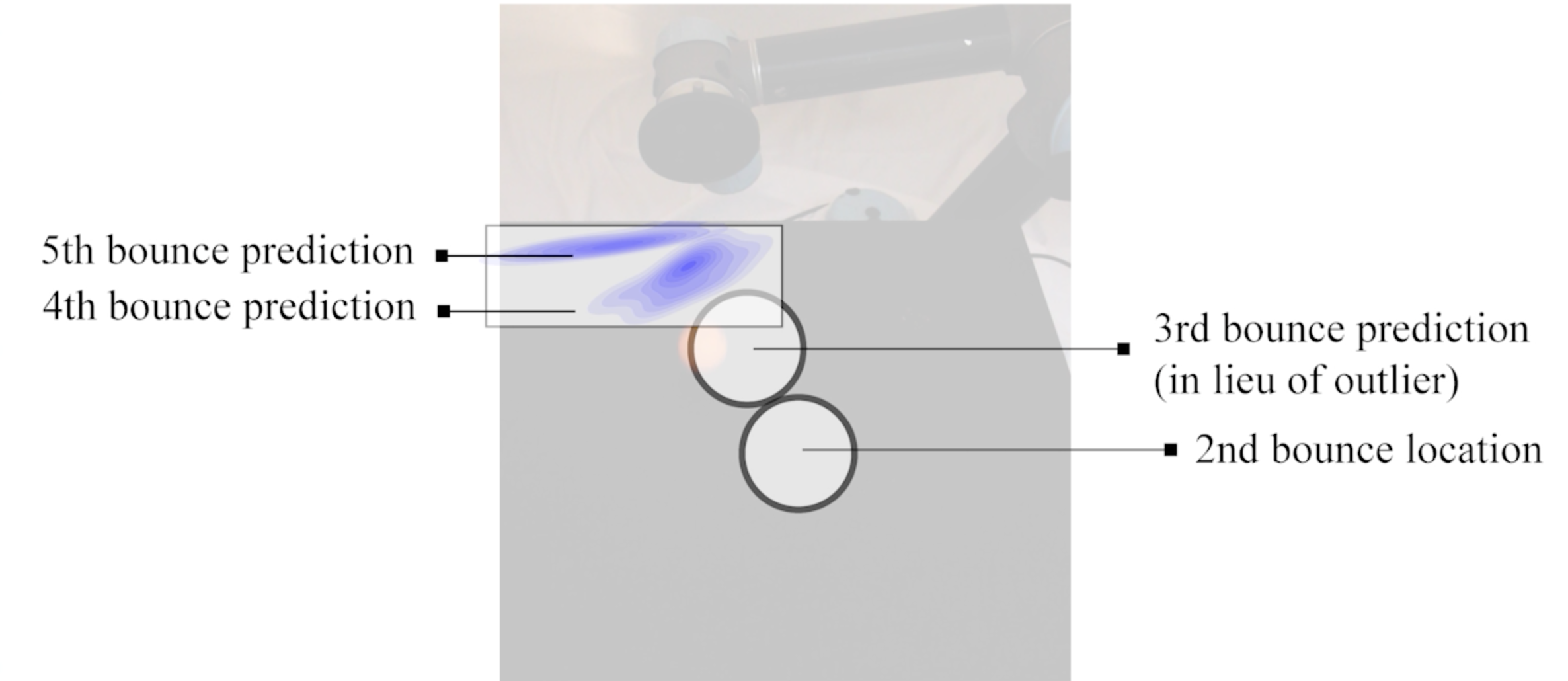}
    \caption{The 3rd \textit{predicted} location, as well as the 4th bounce location distribution, are used to predict the distribution of the 5th bounce.}
    \label{fig:21_10}
    \end{figure}
    
    \clearpage

\begin{figure}[h]
    \centering
    \includegraphics[width=0.7\textwidth]{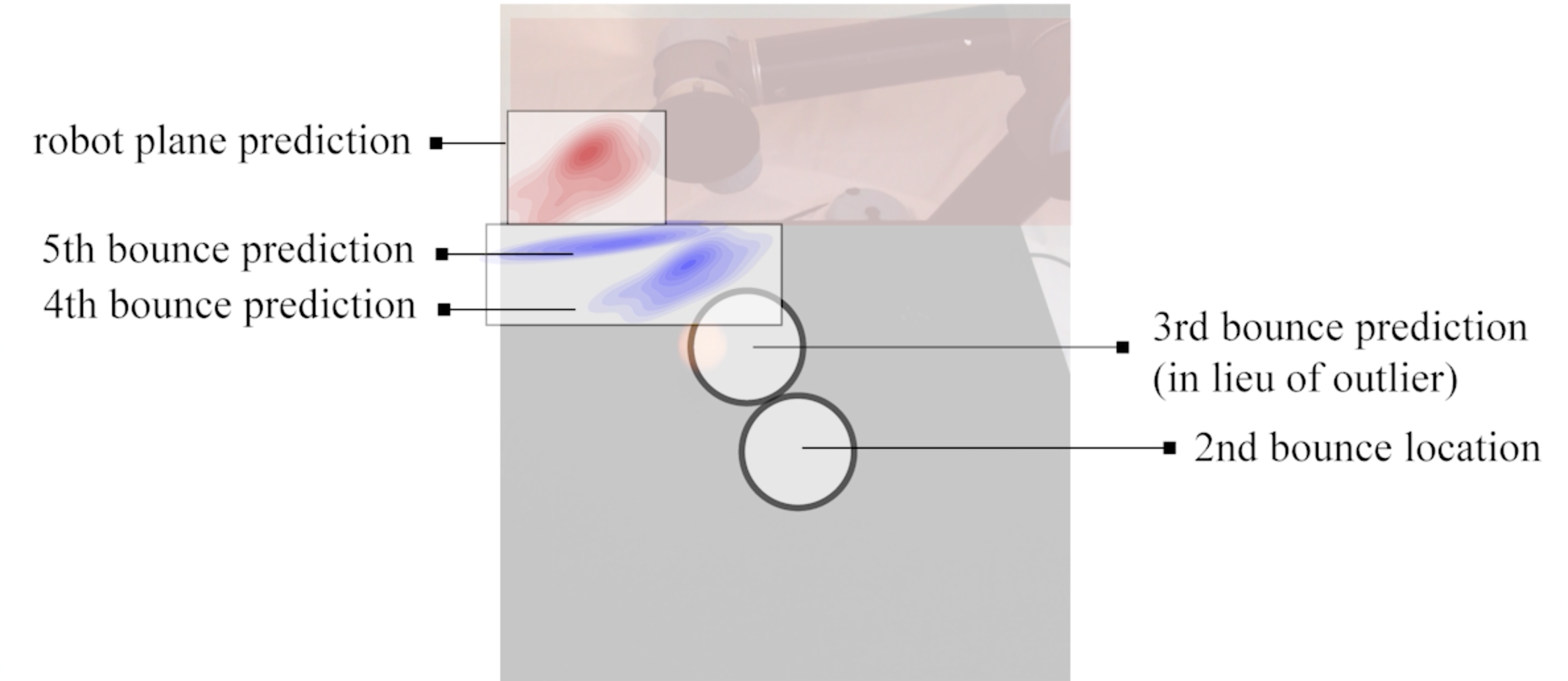}
    \caption{The 4th and 5th bounce location distributions are used to predict the distribution of the robot plane intersections.}
    \label{fig:21_11}
    \end{figure}
    
\begin{figure}[h]
    \centering
    \includegraphics[width=0.7\textwidth]{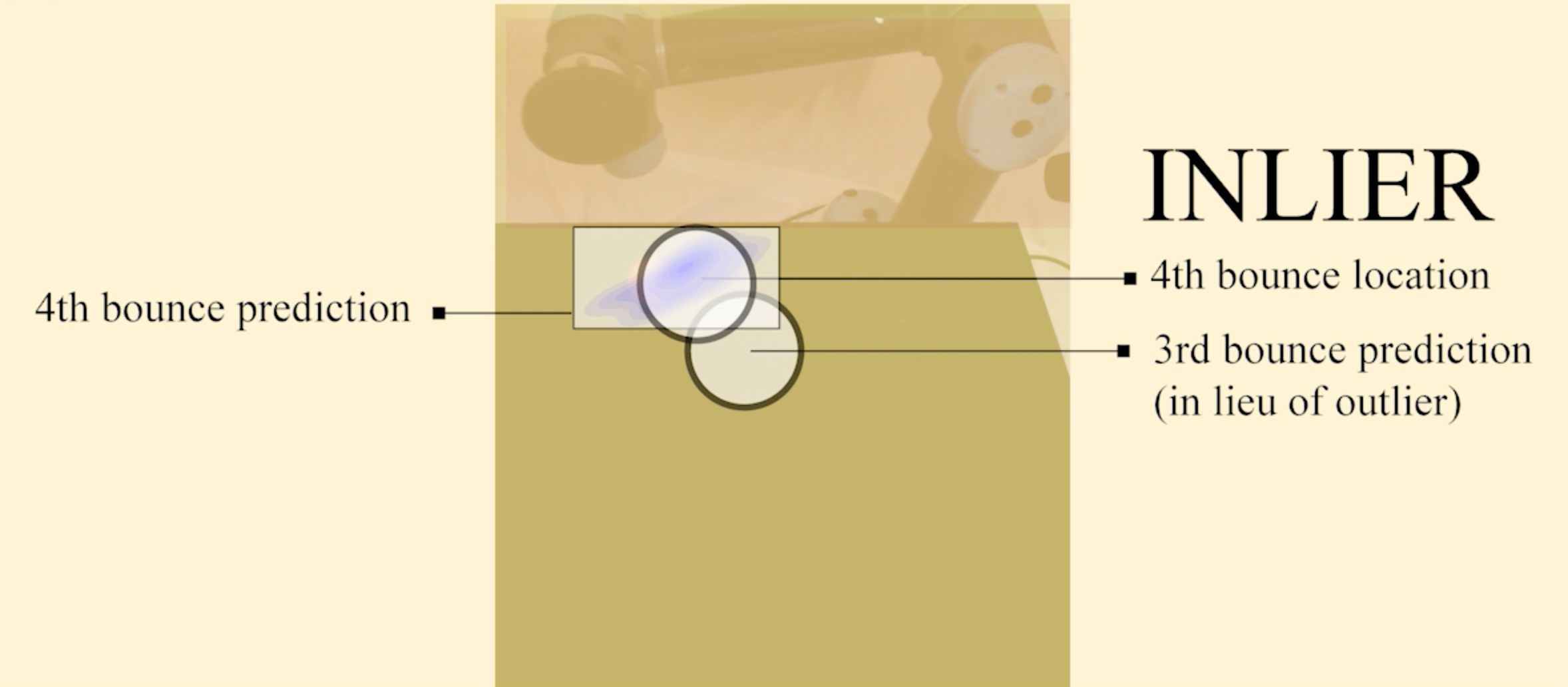}
    \caption{Inlier: the audio-based localization of the 4th bounce lies within the predicted distribution.}
    \label{fig:21_12}
    \end{figure}
    
\begin{figure}[h]
    \centering
    \includegraphics[width=0.7\textwidth]{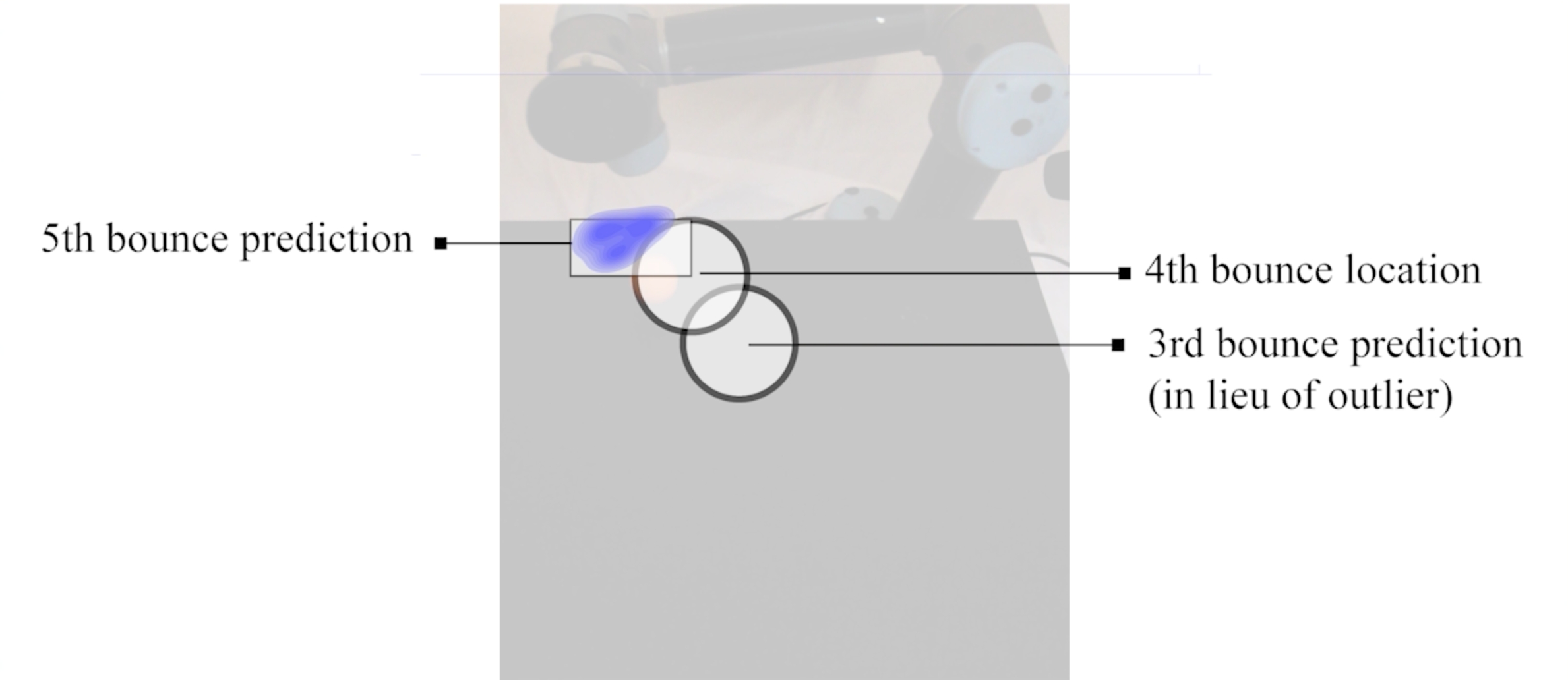}
    \caption{The 3rd \textit{predicted} location, as well as the 4th measured location, are used to predict the distribution of the 5th bounce.}
    \label{fig:21_13}
    \end{figure}
    
    \clearpage
    
\begin{figure}[h]
    \centering
    \includegraphics[width=0.7\textwidth]{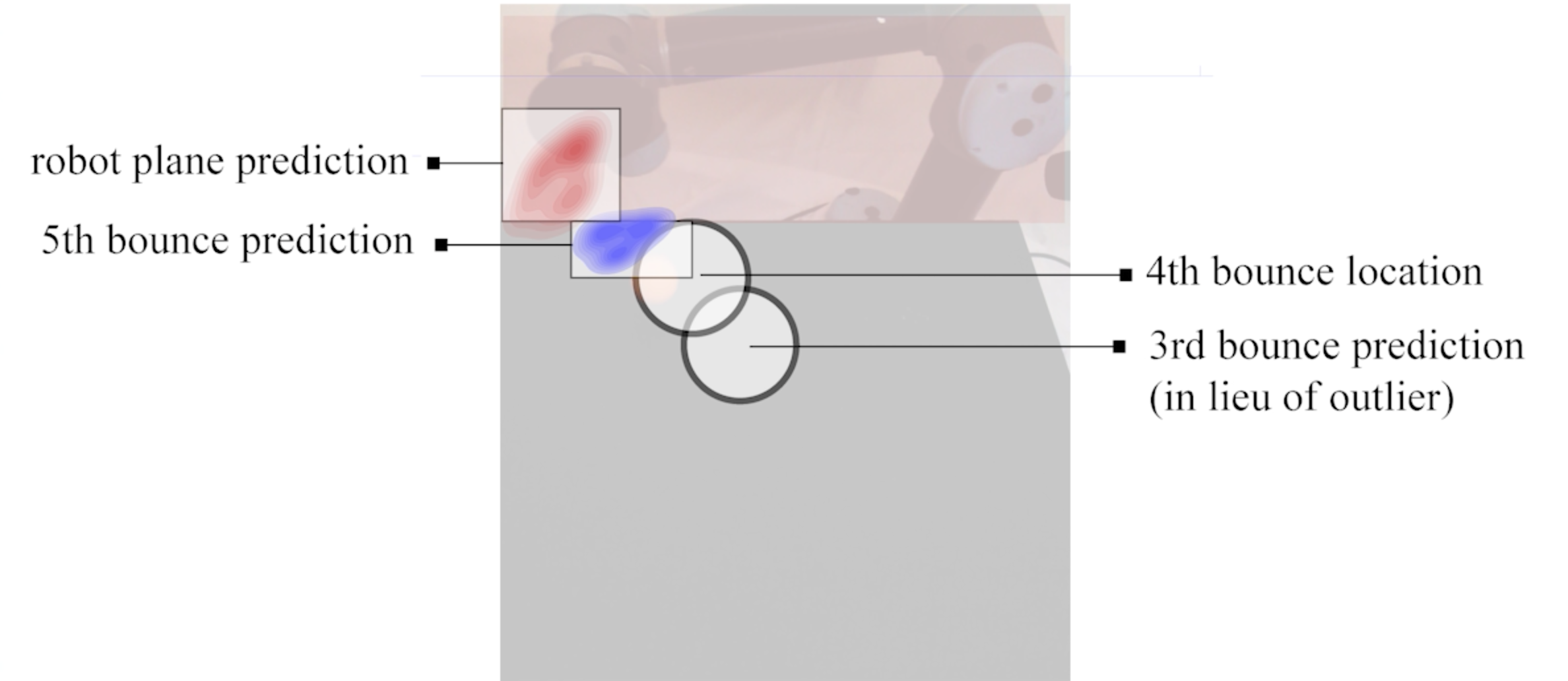}
    \caption{The 4th measured location and the distribution of the 5th bounce are used to predict the distribution of the robot plane intersections.}
    \label{fig:21_14}
    \end{figure}
    
\begin{figure}[h]
    \centering
    \includegraphics[width=0.7\textwidth]{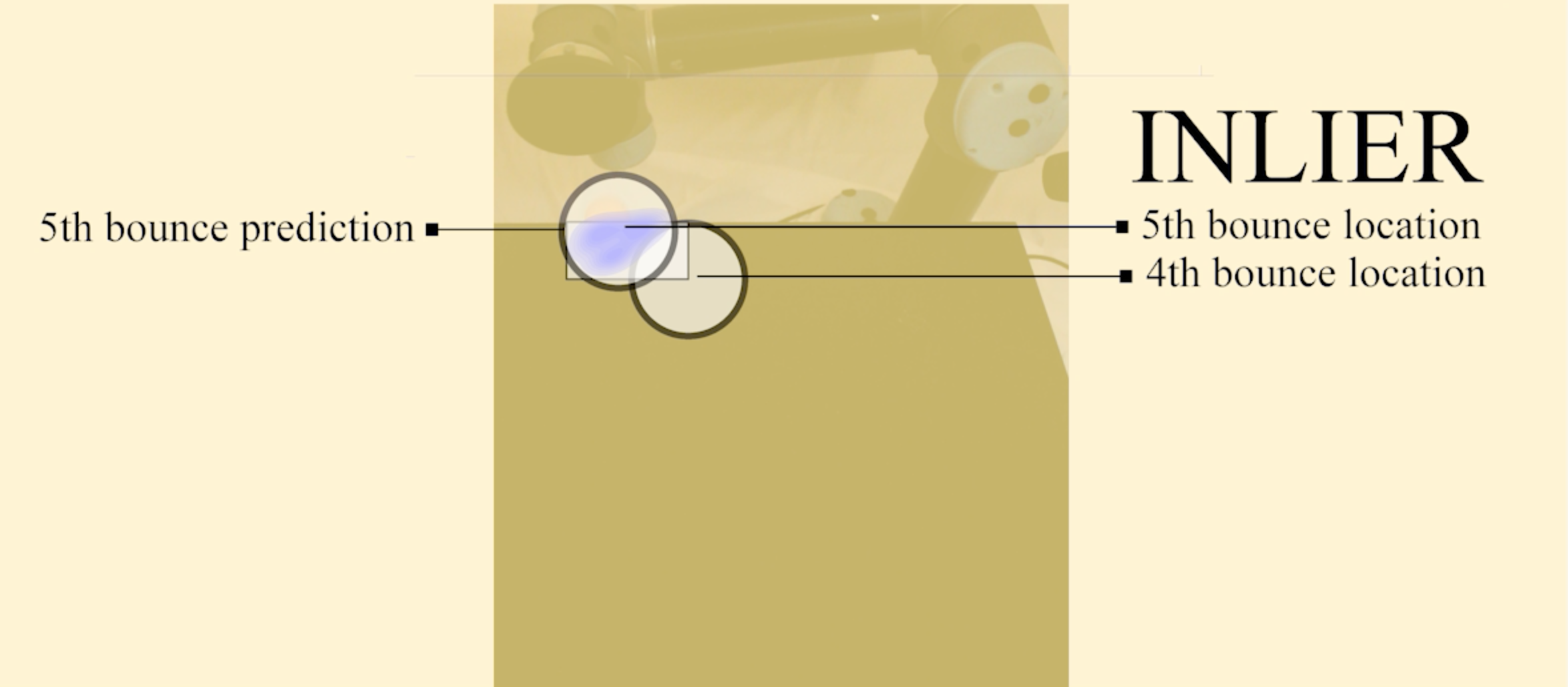}
    \caption{Inlier: the audio-based localization of the 5th bounce lies within the predicted distribution.}
    \label{fig:21_15}
    \end{figure}
    
\begin{figure}[h]
    \centering
    \includegraphics[width=0.7\textwidth]{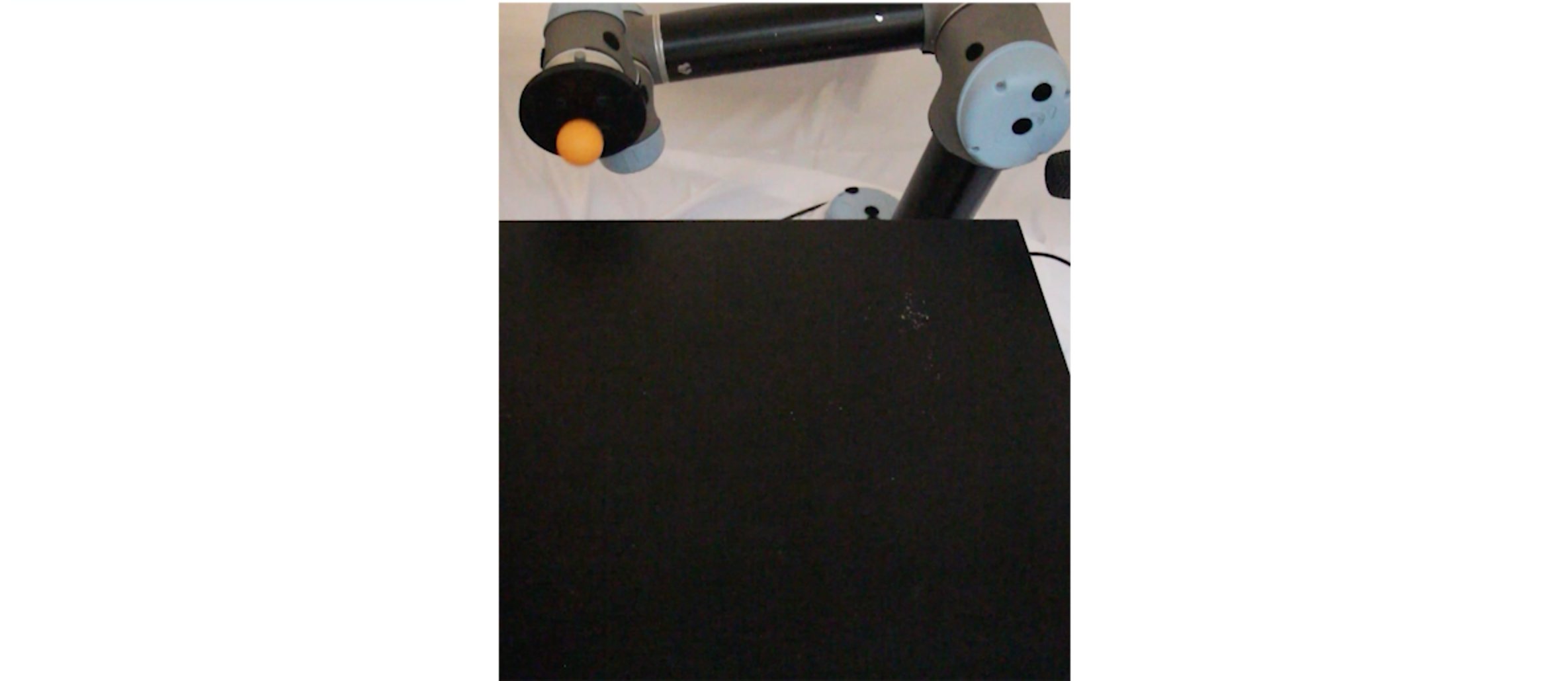}
    \caption{Successful prediction}
    \label{fig:21_17}
    \end{figure}

\end{document}